\documentclass[10pt,journal,compsoc]{IEEEtran}

\usepackage{soul}
\usepackage{url}
\usepackage{ulem}
\usepackage{booktabs}
\usepackage{amsmath}
\usepackage{amsthm} 
\usepackage{amsfonts}  
\usepackage{amssymb}
\usepackage{multicol}

\usepackage[dvipsnames]{xcolor}

\usepackage{multirow}
\usepackage{graphicx}
\usepackage{booktabs}
\usepackage{colortbl}
\usepackage{stfloats}
\usepackage{comment}
\usepackage{makecell}
\usepackage[noend]{algpseudocode}
\usepackage{algorithmicx,algorithm}
\usepackage{pifont}
\usepackage{color}
\usepackage{microtype}
\usepackage{cleveref}
\usepackage[]{footmisc}

\usepackage{soul}
\usepackage{url}
\usepackage{ulem}
\usepackage{booktabs}
\usepackage{amsmath}
\usepackage{amsthm} 
\usepackage{amsfonts}  
\usepackage{amssymb}
\usepackage{multicol}
\usepackage{multirow}
\usepackage{graphicx}
\usepackage{booktabs}
\usepackage{colortbl}
\usepackage{stfloats}
\usepackage{comment}
\usepackage{makecell}
\usepackage[noend]{algpseudocode}
\usepackage{algorithmicx,algorithm}
\usepackage{pifont}
\usepackage{color}
\usepackage{microtype}

\newcommand{\tabincell}[2]{\begin{tabular}{@{}#1@{}}#2\end{tabular}}
\ifCLASSOPTIONcompsoc
  \usepackage[nocompress]{cite}
\else
  \usepackage{cite}
\fi

\begin{document}

\title{Co-guiding for Multi-intent Spoken Language Understanding}

\author{Bowen~Xing and~Ivor~W.~Tsang,~\IEEEmembership{Fellow,~IEEE} 
\IEEEcompsocitemizethanks{\IEEEcompsocthanksitem Bowen Xing is with Australian Artificial Intelligence Institute (AAII), University of Technology Sydney, Australia; CFAR, Agency for Science, Technology and Research, Singapore; and IHPC, Agency for Science, Technology and Research, Singapore.\protect\\
E-mail: bwxing714@gmail.com
\IEEEcompsocthanksitem Ivor Tsang is with CFAR, Agency for Science, Technology and Research, Singapore; 
IHPC, Agency for Science, Technology and Research, Singapore; School of Computer Science and Engineering, Nanyang Technological University; and AAII, University of Technology, Australia. \protect\\
E-mail: ivor\_tsang@ihpc.a-star.edu.sg}
}

\markboth{Journal of \LaTeX\ Class Files,~Vol.~14, No.~8, August~2015}%
{Shell \MakeLowercase{\textit{et al.}}: Bare Demo of IEEEtran.cls for Computer Society Journals}

\IEEEtitleabstractindextext{%
\begin{abstract}
Recent graph-based models for multi-intent SLU have obtained promising results through modeling the guidance from the prediction of intents to the decoding of slot filling.
However, existing methods (1) only model the \textit{unidirectional guidance} from intent to slot, while there are bidirectional inter-correlations between intent and slot; (2) adopt \textit{homogeneous graphs} to model the interactions between the slot semantics nodes and intent label nodes, which limit the performance.
In this paper, we propose a novel model termed Co-guiding Net, which implements a two-stage framework achieving the \textit{mutual guidances} between the two tasks.
In the first stage, the initial estimated labels of both tasks are produced, and then they are leveraged in the second stage to model the mutual guidances.
Specifically, we propose two \textit{heterogeneous graph attention networks} working on the proposed two \textit{heterogeneous semantics-label graphs}, which effectively represent the relations among the semantics nodes and label nodes.
Besides, we further propose Co-guiding-SCL Net, which exploits the single-task and dual-task semantics contrastive relations.
For the first stage, we propose single-task supervised contrastive learning, and for the second stage, we propose co-guiding supervised contrastive learning, which considers the two tasks' mutual guidances in the contrastive learning procedure.
Experiment results on multi-intent SLU show that our model outperforms existing models by a large margin, obtaining a relative improvement of 21.3\% over the previous best model on MixATIS dataset in overall accuracy.
We also evaluate our model on the zero-shot cross-lingual scenario and the results show that our model can relatively improve the state-of-the-art model by 33.5\% on average in terms of overall accuracy for the total 9 languages.
\end{abstract}

}

\maketitle

\IEEEdisplaynontitleabstractindextext

\IEEEpeerreviewmaketitle

\section{Introduction}\label{sec:introduction}
Spoken language understanding (SLU) \cite{slu} is a fundamental task in dialog systems.
Its objective is to capture the comprehensive semantics of user utterances, and it typically includes two subtasks: intent detection and slot filling \cite{idsf}.
Intent detection aims to predict the intention of the user utterance and slot filling aims to extract additional information or constraints expressed in the utterance.

Recently, researchers discovered that these two tasks are closely tied, and a bunch of models \cite{slot-gated,selfgate,cmnet,sfid,qin2019} are proposed to combine the single-intent detection and slot filling in multi-task frameworks to leverage their correlations.

However, in real-world scenarios, a user usually expresses multiple intents in a single utterance.
To this end, \cite{kim2017} begin to tackle the multi-intent detection task and \cite{2019-joint} make the first attempt to jointly model the multiple intent detection and slot filling in a multi-task framework.
\cite{agif} propose an AGIF model to adaptively integrate the fine-grained multi-intent prediction information into the autoregressive decoding process of slot filling via graph attention network (GAT) \cite{gat}.
And \cite{glgin} further propose a non-autoregressive GAT-based model which enhances the interactions between the predicted multiple intents and the slot hidden states, obtaining state-of-the-art results and significant speedup.

\begin{figure}[t]
 \centering
 \includegraphics[width = 0.48\textwidth]{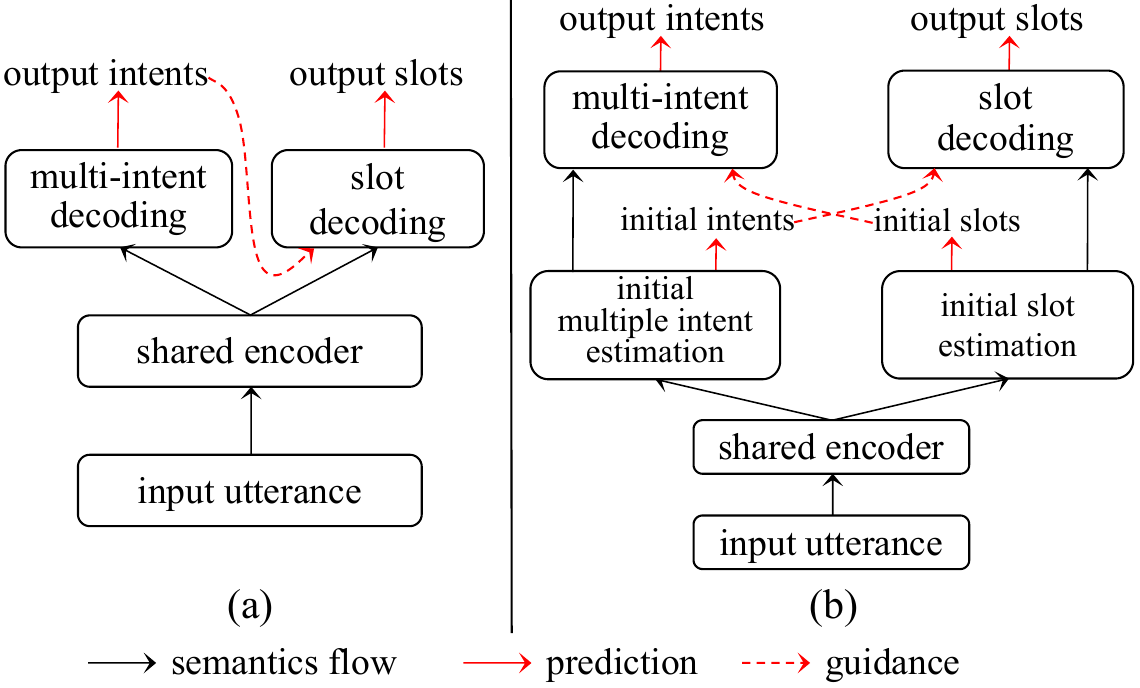}
 \caption{(a) Previous framework which only models the unidirectional guidance from multi-intent predictions to slot filling. (b) Our framework which models the mutual guidances between the two tasks.}
 \label{fig: framework}
\end{figure}

\begin{figure}[t]
 \centering
 \includegraphics[width = 0.45\textwidth]{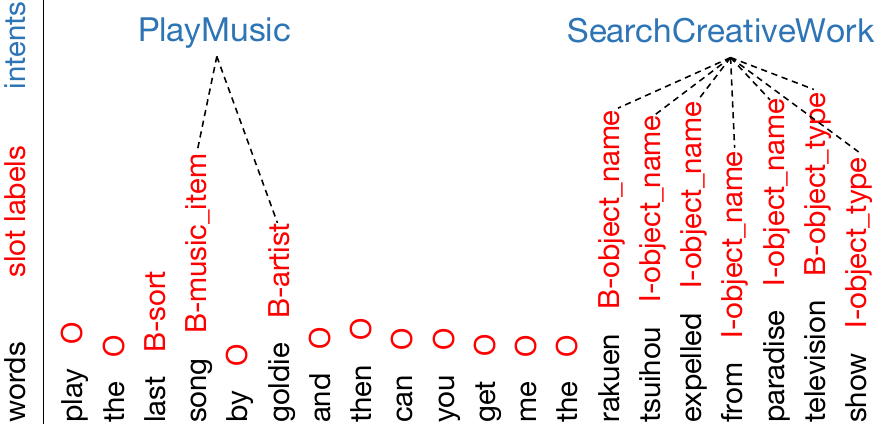}
 \caption{Illustration of the bidirectional interrelations between intent (\textcolor{NavyBlue}{blue}) and slot (\textcolor{red}{red}) labels. The sample is retrieved from MixSNIPS dataset.}
 \label{fig: example}
\end{figure}

Despite the promising progress that existing multi-intent SLU joint models have achieved, we discover that they suffer from two main issues:

(1) \textbf{Ignoring the guidance from slot to intent}. Since previous researchers realized that ``slot labels could depend on the intent'' \cite{2019-joint}, existing models leverage the information of the predicted intents to guide slot filling, as shown in Fig. \ref{fig: framework}(a).
However, they ignore that slot labels can also guide the multi-intent detection task.
Based on our observations, multi-intent detection and slot filling are bidirectionally interrelated and can mutually guide each other.
For example, in Fig \ref{fig: example}, not only the intents can indicate the slots, but also the slots
can infer the intents.
However, in previous works, 
the only guidance that the multiple intent detection task can get from the joint model is sharing the basic semantics with the slot filling task.
As a result, the lack of guidance from slot to intent limits multiple intent detection, and so the joint task.

(2) \textbf{Node and edge ambiguity in the semantics-label graph.} 
\cite{agif,glgin} apply GATs over the constructed graphs to model the interactions among the slot semantics nodes and intent label nodes.
However, their graphs are homogeneous, in which all nodes and edges are treated as the same type.
For a slot semantics node, the information from intent label nodes and other slot semantics nodes play different roles, while the homogeneous graph cannot discriminate their specific contributions, causing ambiguity.
Therefore, the heterogeneous graphs should be designed to represent the relations among the semantic nodes and label nodes to facilitate better interactions.

In this paper, we propose a novel model termed Co-guiding Net to tackle the above two issues.
For the first issue, Co-guiding Net implements a two-stage framework as shown in Fig. \ref{fig: framework} (b).
The first stage produces the initial estimated labels for the two tasks and the second stage leverages the estimated labels as prior label information to allow the two tasks mutually guide each other.
For the second issue, we propose two heterogeneous semantics-label graphs (HSLGs): (1) a slot-to-intent semantics-label graph (S2I-SLG) that effectively represents the relations among the intent semantics nodes and slot label nodes; (2) an intent-to-slot semantics-label graph (I2S-SLG) that effectively represents the relations among the slot semantics nodes and intent label nodes.
Moreover, two heterogeneous graph attention networks (HGATs) are proposed to work on the two proposed graphs for modeling the guidances from slot to intent and intent to slot, respectively.

\begin{figure}[t]
 \centering
 \includegraphics[width = 0.49\textwidth]{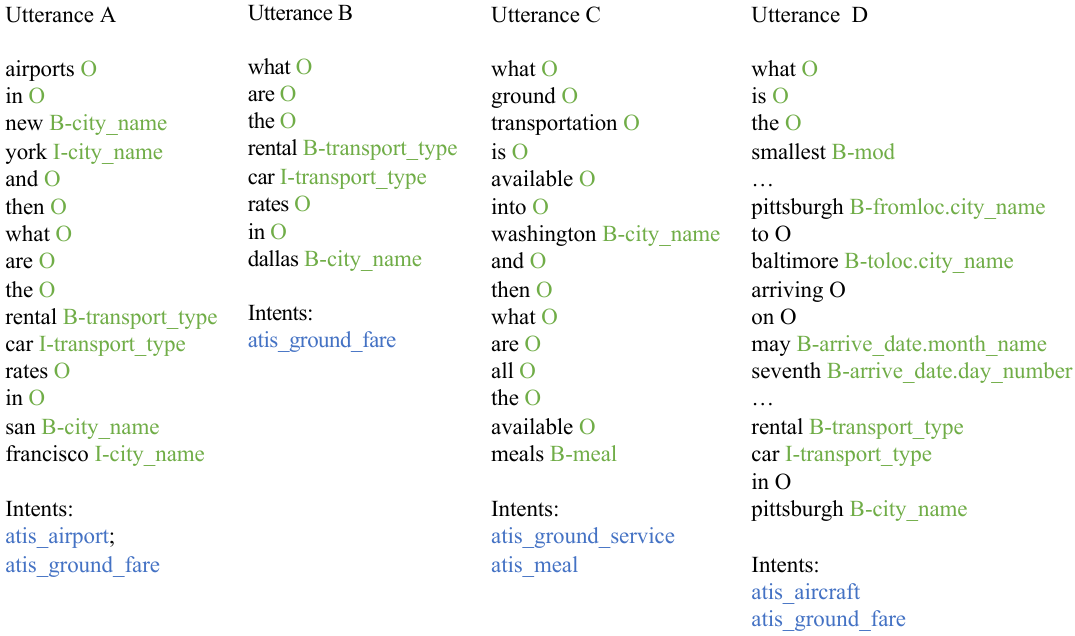}
 \caption{Illustration of some utterances and their intent and slot labels. Intent labels are in \textcolor{NavyBlue}{blue} while slot labels are in \color{LimeGreen}{green}.}
 \label{fig: examples}
\end{figure}

To further leverage the subtle semantic differences among the two tasks' instances, aka semantics contrastive relations, which are ignored by previous studies, we propose Co-guiding-SCL Net, which is based on Co-guiding Net and introduces the supervised contrastive learning to draw together the semantics with the same/similar labels and push apart the semantics with different labels.
In the \textbf{first stage}, since the two tasks are performed individually, we propose two specific single-task supervised contrastive learning mechanisms for multiple intent detection and slot filling, respectively.
Since multiple intent detection is a multi-label classification task, the relationships among instances are not simple positive/negative samples.
To handle the fine-grained correlations among the multi-intent instances, we propose multi-intent supervised contrastive learning that can dynamically assign a fine-grained weight for each instance regarding the similarity of its intents and the anchor's intents.
For slot filling, we adopt the conventional single-label multi-class supervised contrastive learning. 
In the \textbf{second stage}, since the mutual guidances between the two tasks are achieved, there exist dual-task semantics contrastive relations.
As shown in the examples in Fig. \ref{fig: examples}, utterance A and utterance D express similar intents with utterance B (they share the intent label atis\_ground\_fare).
However, we can observe that utterance A and utterance B have more similar sentence-level semantics, while utterance D's semantics is more different.
Utterances A and B mention transporting in some cities, while utterance D contains information about transferring from where to where, the mode, and the date.
This can be reflected in the fact that utterance D has quite different slot labels from utterance A and utterance B.
In the same way, although the word `san' in utterance A, the word `dallas' in utterance B and the word `washington' in utterance C correspond to the same slot label B-city\_name, `san' and `dallas' should have more similar semantics, while `washington' should have more different semantics.
The reason is that `san' and `dallas' have more similar contextual semantics than the contextual semantics of `washington', which can be reflected from the intent labels of utterance A, utterance B and utterance C.
Therefore, intent and slot labels subtly impact the semantics of each other's tasks, which are based on sentence-level and word-level semantics, respectively.
And this subtle indicative information can be leveraged as a supervision signal to benefit the dual-task mutual guidances via leveraging the above dual-task contrastive relations.
Motivated by this, we propose co-guiding supervised contrastive learning to integrate the dual-task correlations in the contrastive learning procedure.
The distances among one task's representations are adjusted regarding not only the own task's contrastive labels but also the guidance from the other task's contrastive labels.
Note that all contrastive learning mechanisms only work in the training process.

The initial version of this work \cite{coguiding} was published on EMNLP 2022 as an oral presentation. 
Its contributions are three-fold:\\
(1) We propose Co-guiding Net, which implements a two-stage framework allowing multiple intent detection and slot filling mutually guide each other.
To the best of our knowledge, this is the first attempt to achieve mutual guidances between the two tasks.\\
(2) We propose two heterogeneous semantics-label graphs as appropriate platforms for the dual-task interactions between semantics nodes and label nodes.
And we propose two heterogeneous graph attention networks to model the mutual guidances between the two tasks.\\
(3) Experiment results on two public multi-intent SLU datasets show that our Co-guiding Net significantly outperforms previous models, and model analysis further verifies the advantages of our model.

In this paper, we significantly extend our work from the previous version in the following aspects:\\
(1) We propose Co-guiding-SCL Net, which augments Co-guiding Net with  supervised contrastive learning mechanisms to further capture the single-task and dual-task semantics contrastive relations among the samples.\\
(2) For the first stage, we propose the single-task supervised contrastive learning mechanism for both tasks.
For multiple intent detection, we propose a novel multi-intent supervised contrastive learning mechanism to capture the dynamic and fine-grained correlations between the multi-intent instances.\\
(3) For the second stage, we propose co-guiding supervised contrastive learning, which can capture the fine-trained dual-task semantics contrastive correlations by jointly considering both tasks' labels as the supervision signal to perform supervised contrastive learning for each task.\\
(4) We conduct extensive experiments on the public multi-intent SLU datasets.
Except for LSTM, we also evaluate our model on several pre-trained language model (PTLM) encoders.
The experimental results show that our model can achieve significant and consistent improvements over stage-of-the-art models.
And the model analysis further verifies the advantages of our proposed dual-task supervised contrastive learning mechanisms.\\
(5) We also evaluate our model on the zero-shot cross-lingual multi-intent SLU task, which has never been explored.
The experimental results show that our model can significantly improve the existing best-performing model on the average overall accuracy of the total 9 languages.

The remainder of this paper is organized as follows. In Section 2, we summarize the related works of Spoken Language Understanding, Graph Neural Networks for NLP and Contrastive Learning for NLP.
And the differences between our method and previous studies are highlighted. 
Section 3 elaborates on the details of Co-guiding Net. 
Section 4 depicts the proposed supervised contrastive learning mechanisms in Co-guiding-SCL Net. 
Experimental results are reported and analyzed in Section 5.
Note that the task definition of zero-shot cross-lingual multiple intent detection and slot filling as well as the experiments on this task are introduced in Section \ref{sec: zero-shot}. 
Finally, the conclusion of this work and some prospective future directions are provided in Section 6.

\section{Related Work}\label{sec: relatedwork}
\subsection{Spoken Language Understanding}
The correlations between intent detection and slot filling have been widely recognized.
To leverage them, a group of models \cite{ijcai2016joint,hakkani2016multi,slot-gated,selfgate,sfid,cmnet,qin2019,jointcap,slotrefine,qin2021icassp,ni2021recent} are proposed to tackle the joint task of intent detection and slot filling in a multi-task manner.
However, the intent detection modules in the above models can only handle the utterances expressing a single intent, which may not be practical in real-world scenarios, where there are usually multi-intent utterances.

To this end, \cite{kim2017} propose a multi-intent SLU model, and \cite{2019-joint} propose the first model to jointly model the tasks of multiple intent detection and slot filling via a slot-gate mechanism.
Furthermore, as graph neural networks have been widely utilized in various tasks \cite{cao-etal-2019-multi,rgat,shi-etal-2021-transfernet,kagrmn,darer,dignet}, they have been leveraged to model the correlations between intent and slot.
\cite{agif} propose an adaptive graph-interactive framework to introduce the fine-grained multiple intent information into slot filling achieved by GATs.
More recently, \cite{glgin} propose another GAT-based model, which includes a non-autoregressive slot decoder conducting parallel decoding for slot filling and achieves the state-of-the-art performance.

Our work also tackles the joint task of multiple intent detection and slot filling.
Existing methods only model the one-way guidance from multiple intent detection to slot filling.
Besides, they adopt homogeneous graphs and vanilla GATs to achieve the interactions between the predicted intents and slot semantics.
Different from previous works, we (1) achieve the mutual guidances between the two tasks; (2)  propose the heterogeneous semantics-label graphs to represent the dependencies among the semantics and predicted labels; (3) we propose the Heterogeneous Graph Attention Network to model the semantics-label interactions on the heterogeneous semantics-label graphs; (4) we propose a group of supervised contrastive learning mechanisms to further capture the high-level semantic structures and fine-grained dual-task correlations.

\subsection{Graph Neural Networks for NLP}
In recent years, graph neural networks have been widely adopted in various NLP tasks.
Some works \cite{asgcn,rgat,kagrmn,neuralsubgraph} leverage the GAT and GCN to encode syntactic information for target sentiment classification.
CGR-Net \cite{cgrnet} models the interactions among the emotion-cause pair extraction and the two subtasks through the multi-task relational graph.
In dialog understanding, DARER \cite{darer,darerpami} applies the relational graph convolutional network \cite{rgcn} over the constructed speaker-aware temporal graph and dual-task temporal graph to capture the relational temporal information.
In spoken language understanding, AGIF \cite{agif} and GL-GIN \cite{glgin} leverage graph structures and GNNs to model the intent-slot correlation.
ReLa-Net \cite{rela-net} adopts a heterogeneous label graph to model the dual-task label dependencies.
In this paper, we propose two heterogeneous graphs to provide an appropriate platform for dual-task semantics-label interactions, which is achieved by our proposed heterogeneous attention networks.

\subsection{Contrastive Learning for NLP}
Contrastive learning has been leveraged to improve semantic representations in different NLP tasks \cite{zhang-etal-2021-pairwise,knn-intentcl,wang2022incorporating}.
In natural language inference,  pairwise supervised CL \cite{zhang-etal-2021-pairwise} propose to utilize high-level categorical concept encoding to bridge semantic entailment and contradiction understanding.
Hierarchy-guided contrastive learning \cite{wang2022incorporating} is proposed to directly incorporate the hierarchy into the text encoder for hierarchical text classification.
GL-CLEF \cite{gl-clef} performs unsupervised contrastive learning to achieve the cross-lingual semantics alignment to improve zero-shot cross-lingual intent detection and slot filling.
In this paper, we propose single-task supervised contrastive learning to further capture the single-task semantics contrastive relations in the first stage.
And we propose co-guiding supervised contrastive learning to capture the dual-task semantics contrastive relations in the second stage.
\section{Co-guiding} \label{sec: coguid}
\textbf{Problem Definition.}
Given a input utterance denoted as $U=\{u_i\}^n_1$,
 multiple intent detection can be formulated as a multi-label classification task that outputs multiple intent labels corresponding to the input utterance.
And slot filling is a sequence labeling task that maps each $u_i$ into a slot label.

Next, before diving into the details of Co-guiding Net's architecture, we first introduce the construction of the two heterogeneous graphs.

\subsection{Graph Construction}
\subsubsection{Slot-to-Intent Semantics-Label Graph}
\begin{figure}[t]
 \centering
 \includegraphics[width = 0.4\textwidth]{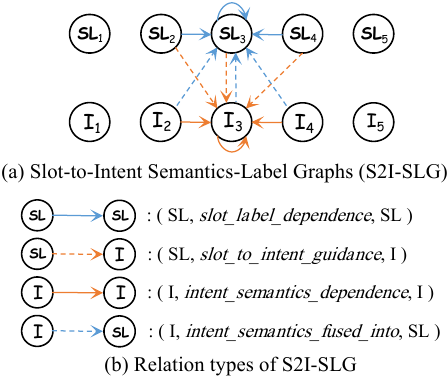}
 \caption{The illustration of S2I-SLG and its relation types. w.l.o.g, only the edges directed into SL$_3$ and I$_3$ are shown, and the local window size is 1.}
 \label{fig: S2I-SLG}
\end{figure}

To provide an appropriate platform for modeling the guidance from the estimated slot labels to multiple intent detection, we design a slot-to-intent semantics-label graph (S2I-SLG), which represents the relations among the semantics of multiple intent detection and the estimated slot labels.
S2I-SLG is a heterogeneous graph and an example is shown in Fig. \ref{fig: S2I-SLG} (a).
It contains two types of nodes: intent semantics nodes\footnote{Each word corresponds to a semantics node. In LSTM setting, semantics node representation is each word's hidden state. In PTLM setting, semantics node representation is each word's first token's hidden state.}
(e.g., I$_1$, ..., I$_5$) and \textbf{s}lot \textbf{l}abel (SL) nodes (e.g., SL$_1$, ..., SL$_5$).
And there are four types of edges in S2I-SLG, as shown in Fig. \ref{fig: S2I-SLG} (b).
Each edge type corresponds to an individual kind of information aggregation on the graph.

Mathematically, the S2I-SLG can be denoted as $\mathcal{G}_{s2i}=\left(\mathcal{V}_{s2i}, \mathcal{E}_{s2i}, \mathcal{A}_{s2i}, \mathcal{R}_{s2i}\right)$, in which $\mathcal{V}_{s2i}$ is the set of all nodes, $\mathcal{E}_{s2i}$ is the set of all edges, $\mathcal{A}_{s2i}$ is the set of two node types and $\mathcal{R}_{s2i}$ is the set of four edge types.
Each node $v_{s2i}$ and each edge $e_{s2i}$ are associated with their type mapping functions $\tau(v_{s2i}): \mathcal{V}_{s2i} \rightarrow \mathcal{A}_{s2i}$ and $\phi(e_{s2i}): \mathcal{E}_{s2i} \rightarrow \mathcal{R}_{s2i}$.
For instance, in Fig. \ref{fig: S2I-SLG}, the SL$_2$ node belongs to $\mathcal{V}_{s2i}$, while its node type SL belongs to $\mathcal{A}_{s2i}$; the edge from SL$_2$ to I$_3$ belongs to $\mathcal{E}_{s2i}$, while its edge type \textit{slot\_to\_intent\_guidance} belongs to $\mathcal{R}_{s2i}$.
Besides, edges in S2I-SLG are based on local connections.
For example, node I$_i$ is connected to $\{\text{I}_{i-w}, ..., \text{I}_{i+w}\}$ and $\{\text{SL}_{i-w}, ..., \text{SL}_{i+w}\}$, where $w$ is a hyper-parameter of the local window size.

\subsubsection{Intent-to-Slot Semantics-Label Graph}

\begin{figure}[t]
 \centering
 \includegraphics[width = 0.41\textwidth]{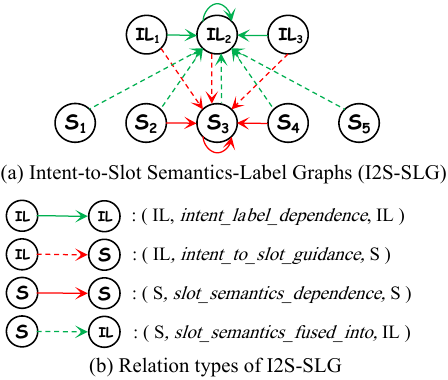}
 \caption{The illustration of I2G-SLG and its relation types. w.l.o.g, only the edges directed into IL$_3$ and S$_3$ are shown, and the local window size is 1.}
 \label{fig: I2S-SLG}
\end{figure}

To present a platform for accommodating the guidance from the estimated intent labels to slot filling, we design an intent-to-slot semantics-label graph (I2S-SLG) that represents the relations among the slot semantics nodes and the intent label nodes.
I2S-SLG is also a heterogeneous graph and an example is shown in Fig. \ref{fig: I2S-SLG} (a).
It contains two types of nodes: slot semantics nodes (e.g., S$_1$, ..., S$_5$) and \textbf{i}ntent \textbf{l}abel (IL) nodes (e.g., IL$_1$, ..., IL$_5$).
And Fig. \ref{fig: I2S-SLG} (b) shows the four edge types.
Each edge type corresponds to an individual kind of information aggregation on the graph.

Mathematically, the I2S-SLG can be denoted as $\mathcal{G}_{i2s}=\left(\mathcal{V}_{i2s}, \mathcal{E}_{i2s}, \mathcal{A}_{i2s}, \mathcal{R}_{i2s}\right)$.
Each node $v_{i2s}$ and each edge $e_{i2s}$ are associated with their type mapping functions $\tau(v_{i2s})$
and $\phi(e_{i2s})$. 
The connections in I2S-SLG are a little different from S2I-SLG.
Since intents are sentence-level, each IL node is globally connected with all nodes.
For S$_i$ node, it is connected to $\{\text{S}_{i-w}, ..., \text{S}_{i+w}\}$ and $\{\text{IL}_1, ..., \text{IL}_m\}$, where $w$ is the local window size and $m$ is the number of estimated intents.

\subsection{Model Architecture}

\begin{figure*}[t]
 \centering
 \includegraphics[width = 0.9\textwidth]{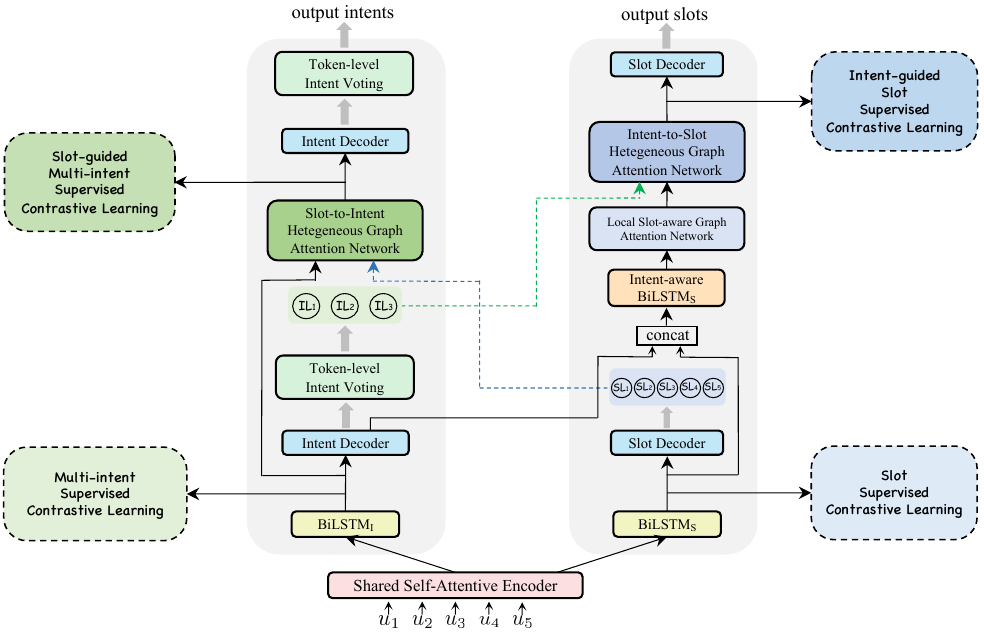}
 \caption{The architecture of Co-guiding Net and Co-guiding-SCL Net. Dashed boxes denote the contrastive learning modules included in Co-guiding-SCL Net while not in Co-guiding Net. Dashed lines denote that the contrastive learning modules only work in the training procedure. Each HGAT is triggered by its own task's semantics and the counterpart's predicted labels.
 The green and blue dashed arrow lines denote the projected label representations from the predicted intents and slots, respectively. The green solid arrow line denotes the intent distribution generated by the Intent Decoder at the first stage.}
 \label{fig: model}
\end{figure*}

In this section, we introduce the details of our Co-guiding Net, whose architecture is shown in Fig.\ref{fig: model}.
\subsubsection{Shared Self-Attentive Encoder}
Following \cite{agif,glgin}, we adopt a shared self-attentive encoder to produce the initial hidden states containing the basic semantics.
It includes a BiLSTM and a self-attention module.
BiLSTM captures the temporal dependencies: 
 \begin{equation}
h_i = \operatorname{BiLSTM} \big(x_i, h_{i-1}, h_{i+1}\big)
\end{equation}
where $x_i$ is the word vector of $u_i$.
Now we obtain the context-sensitive hidden states $\boldsymbol{\hat{H}}=\{\hat{h_i}\}_1^n$.

Self-attention captures the global dependencies: 
\begin{equation}
\boldsymbol{H'}=\operatorname{softmax}\left(\frac{\boldsymbol{Q} \boldsymbol{K}^{\top}}{\sqrt{d_{k}}}\right) \boldsymbol{V}
\end{equation}
where $\boldsymbol{H'}$ is the global contextual hidden states output by self-attention; $\boldsymbol{Q}, \boldsymbol{K}$ and $\boldsymbol{V}$ are matrices obtained by applying different linear projections on the input utterance word vector matrix. 

Then we concatenate the output of BiLSTM and self-attention to form the output of the shared self-attentive encoder:
$\boldsymbol{H}=\boldsymbol{\hat{H}} \| \boldsymbol{H'}$, where $\boldsymbol{H}=\{h_i\}_1^n$ and $\|$ denotes concatenation operation.

\subsubsection{Initial Estimation}
\textbf{Multiple Intent Detection.}
To obtain the task-specific features for multiple intent detection, we apply a BiLSTM layer over $\boldsymbol{H}$:
 \begin{equation}
h^{[I,0]}_i = \operatorname{BiLSTM}_{\text{I}} \left(h_i, h^{[I,0]}_{i-1}, h^{[I,0]}_{i+1}\right)
\end{equation}

Following \cite{agif,glgin}, we conduct token-level multi-intent detection.
Each $h^{[I,0]}_i$ is fed into the intent decoder. 
Specifically, the intent label distributions of the $i$-th word are obtained by:
 \begin{equation}
y^{[I,0]}_{i}=\operatorname{sigmoid}\left(\boldsymbol{W}^1_{I}\left(\sigma(\boldsymbol{W}^2_{I} \boldsymbol{h}_{i}^{[I,0]}\!+\!\boldsymbol{b}^2_{I})\right)\!+\!\boldsymbol{b}^1_{I}\right)
\end{equation}
where $\sigma$ denotes the non-linear activation function; $W_*$ and $b_*$ are model parameters.

Then the estimated sentence-level intent labels $\{\text{IL}_1, ..., \text{IL}_m\}$ are obtained by the token-level intent voting \cite{glgin}.

\noindent\textbf{Slot Filling.}
\cite{glgin} propose a non-autoregressive paradigm for slot filling decoding, which achieves significant speedup.
In this paper, we also conduct parallel slot filling decoding.

We first apply a BiLSTM over $\boldsymbol{H}$ to obtain the task-specific features for slot filling:
\begin{equation}
h^{[S,0]}_i = \operatorname{BiLSTM}_{\text{S}} (h_i, h^{[S,0]}_{i-1}, h^{[S,0]}_{i+1})
\end{equation}
Then use a softmax classifier to generate the slot label distribution for each word:
\begin{equation}
y_{i}^{[S,0]}=\operatorname{softmax} \left(\boldsymbol{W}^1_{S}\left(\sigma (\boldsymbol{W}^2_{S} \boldsymbol{h}_{i}^{[S,0]}\!+\!\boldsymbol{b}^2_{S})\right)\!+\!\boldsymbol{b}^1_{S}\right)
\end{equation}
And the estimated slot label for each word is obtained by $\text{SL}_i = \operatorname{arg\ max}(y_{i}^{[S,0]})$.

\subsubsection{Heterogeneous Graph Attention Network}
State-of-the-art models \cite{agif,glgin} use a homogeneous graph to connect the semantic nodes of slot filling and the intent label nodes.
And GAT \cite{gat} is adopted to achieve information aggregation.
In Sec. \ref{sec:introduction}, we propose that this manner cannot effectively learn the interactions between one task's semantics and the estimated labels of the other task.
To tackle this issue, we propose two heterogeneous graphs (S2I-SLG and I2S-SLG) to effectively represent the relations among the semantic nodes and label nodes.
To model the interactions between semantics and labels on the proposed graphs, we propose a Heterogeneous Graph Attention Network (HGAT).
When aggregating the information into a node, HGAT can discriminate the specific information from different types of nodes along different relations.
And two HGATs (S2I-HGAT and I2S-HGAT) are applied on S2I-SLG and I2S-SLG, respectively.
Specifically, 
 S2I-HGAT can be formulated as follows: 
\begin{equation}
 \begin{split}
&h^{l+1}_i \!= \! \mathop{\|}\limits_{k=1}^K\sigma\!\left(\sum_{j\in \mathcal{N}_{s2i}^{i}} \!\!\! W_{s2i}^{[r,k,1]} \alpha_{ij}^{[r,k]} h^{l}_j \!\right)\!,  r = \phi\!\left(e^{[j,i]}_{s2i}\right)\\
&\alpha_{ij}^{[r,k]}\! = \!\frac{\operatorname{exp} \!\left( \! \left( \!W_{s2i}^{[r,k,2]} h^{l}_i \! \right) \! \left( \!W_{s2i}^{[r,k,3]} h^{l}_j \!\right)^\mathsf{T} \!\!/\! \sqrt d  \right)}{\sum\limits_{u\in\mathcal{N}_{s2i}^{r,i}} \! \operatorname{exp} \! \left( \! \left( \!W_{s2i}^{[r,k,2]} h^{l}_i \!\right) \! \left( \!W_{s2i}^{[r,k,3]} h^{l}_u \!\right)^\mathsf{T} \!\!\!/ \!\sqrt d \right)}
\end{split}\label{eq: hgat}
\end{equation}
where $K$ denotes the total head number; $\mathcal{N}_{s2i}^{i}$ denotes the set of incoming neighbors of node $i$ on S2I-SLG; $W_{s2i}^{[r,k,*]}$ are weight matrices of edge type $r$ on the $k$-th head; $e^{[j,i]}_{s2i}$ denotes the edge from node $j$ to node $i$ on S2I-SLG; $\mathcal{N}_{s2i}^{r,i}$ denotes the nodes connected to node $i$ with $r$-type edges on S2I-SLG; $d$ is the dimension of node hidden state.

I2S-HGAT can be derived like Eq. \ref{eq: hgat}.
\subsubsection{Intent Decoding with Slot Guidance}
In the first stage, we obtain the initial intent features $H^{[I,0]}=\{h^{I,0}_i\}_i^n$ and the initial estimated slot labels sequence $\{\text{SL}_1, ..., \text{SL}_n\}$.
Now we project the slot labels into vector form using the slot label embedding matrix, obtaining $E_{sl}=\{e^1_{sl}, ..., e^n_{sl}\}$.

Then we feed $H^{[I,0]}$ and $E_{sl}$ into S2I-HGAT to model their interactions, allowing the estimated slot label information to guide the intent decoding:
\begin{equation}
 H^{[I,L]}=\operatorname{S2I-HGAT}\left([H^{[I,0]}, E_{sl}],\mathcal{G}_{s2i}, \theta_I\right)
\end{equation}
where $[H^{[I,0]}, E_{sl}]$ denotes the input node representation; $\theta_I$ denotes S2I-HGAT's parameters.
$L$ denotes the total layer number.

Finally, $H^{[I,L]}$ is fed to intent decoder, producing the intent label distributions for the utterance words: $Y^{[I,1]}=\{y^{[I,1]}_i, ..., y_n^{[I,1]}\}$.
And the final output sentence-level intents are obtained via applying token-level intent voting over $Y^{[I,1]}$.
\subsubsection{Slot Decoding with Intent Guidance}
\noindent\textbf{Intent-aware BiLSTM.} Since the B-I-O tags of slot labels have temporal dependencies, we use an intent-aware BiLSTM to model the temporal dependencies among slot hidden states with the guidance of estimated intents:
\begin{equation}
 \tilde{h}_i^{[S,0]}=\operatorname{BiLSTM}(y_i^{[I,0]}\|h_i^{[S,0]}, \tilde{h}_{i-1}^{[S,0]}, \tilde{h}_{i+1}^{[S,0]})
\end{equation}

\noindent\textbf{I2S-HGAT.} We first project the estimated intent labels $\{\text{IL}_j\}_1^m$ into vectors using the intent label embedding matrix, obtaining $E_{il}=\{e^1_{il}, ..., e^m_{il}\}$.
Then we feed $\tilde{H}^{S}$ and $E_{il}$ into I2S-HGAT to model their interactions, allowing the estimated intent label information to guide the slot decoding:
\begin{equation}
 H^{[S,L]}=\operatorname{I2S-HGAT}\left([\tilde{H}^{S}, E_{il}],\mathcal{G}_{i2s}, \theta_S\right)
\end{equation}
where $[\tilde{H}^{[S]}, E_{il}]$ denotes the input node representation; $\theta_S$ denotes I2S-HGAT's parameters.

Finally, $H^{[S,L]}$ is fed to slot decoder, producing the slot label distributions for each word: $Y^{[S,1]}=\{y^{[S,1]}_i, ..., y_n^{[S,1]}\}$.
And the final output slot labels are obtained by applying $\operatorname{arg\ max}$ over $Y^{[S,1]}$.

\begin{figure*}[t]
 \centering
 \includegraphics[width = \textwidth]{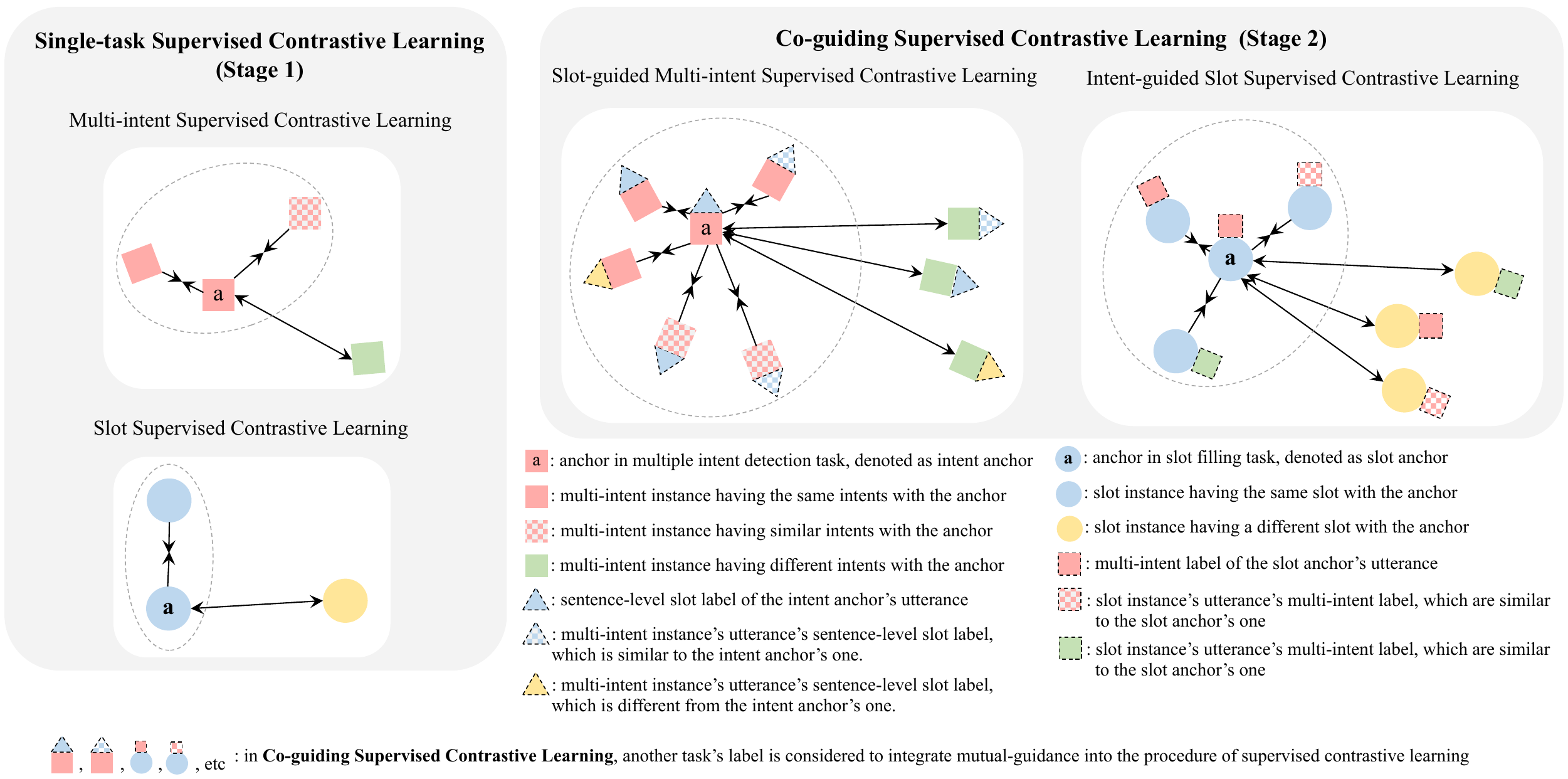}
 \caption{Conceptual illustration of single-task supervised contrastive learning, which is performed in stage 1, and co-guiding supervised contrastive learning, which is performed in stage 2.}
 \label{fig: scl_model}
\end{figure*}

\subsection{Training Objective}
\subsubsection{Loss Function}
The loss function for multiple intent detection is:
\begin{equation}
 \begin{split}
\mathrm{CE}(\hat{y}, y)&=\hat{y} \log (y)+(1-\hat{y}) \log (1-y)\\
\mathcal{L}_{I} &=\sum_{t=0}^1\sum_{i=1}^{n} \sum_{j=1}^{N_{I}} \mathrm{CE}\left(\hat{y}_{i}^{I}[j], y_{i}^{[I,t]}[j]\right)
\end{split}
\end{equation}
And the loss function for slot filling is:
\begin{equation}
\mathcal{L}_{S} = \sum_{t=0}^1\sum_{i=1}^{n} \sum_{j=1}^{N_{S}} \hat{y}_{i}^{S}[j] \log \left(y_{i}^{[S,t]}[j]\right)
\end{equation}
where ${N}_I$ and ${N}_S$ denote the total numbers of intent labels and slot labels; $\hat{y}_{i}^{I}$ and $\hat{y}_{i}^{S}$ denote the ground-truth intent labels and slot labels. 
\subsubsection{Margin Penalty}
The core of our model is to let the two tasks mutually guide each other.
Intuitively, the predictions in the second stage should be better than those in the first stage.
To force our model to obey this rule, we design a margin penalty ($\mathcal{L}^{mp}$) for each task, whose aim is to improve the probabilities of the correct labels. Specifically, $\mathcal{L}_I^{mp}$ and $\mathcal{L}_S^{mp}$ are formulated as:
\begin{equation} 
 \begin{split}
&\mathcal{L}_I^{mp} \!=\! \sum_{i=1}^n \sum_{j=1}^{{N}_I} \hat{y}_{i}^{I}[j]\ \operatorname{max} \left(0, y_i^{[I,0]}[j]-y_i^{[I,1]}[j]\right)\\
&\mathcal{L}_S^{mp} \!=\! \sum_{i=1}^{n} \sum_{j=1}^{N_{S}} \hat{y}_{i}^{S}[j] \operatorname{max}  \left(0, y_i^{[S,0]}[j]-y_i^{[S,1]}[j]\right)
\end{split}
\end{equation}
\subsubsection{Model Training}
The training objective $\mathcal{L}$ is the weighted sum of loss functions and margin regularizations of the two tasks: 
\begin{equation}
\mathcal{L} = \gamma\left(\mathcal{L}_I + \beta_I\mathcal{L}_I^{mp}\right) + \left(1-\gamma\right)\left(\mathcal{L}_S + \beta_S \mathcal{L}_S^{mp}\right) \label{eq: final objective}
\end{equation}
where $\gamma$ is the coefficient balancing the two tasks; $\beta_I$ and $\beta_S$ are the coefficients of the margin regularization for the two tasks.

\section{Co-guiding-SCL Net} \label{sec: scl}
Existing methods suffer from three issues: 1) ignoring the guidance from slot to intent; 2) Node and edge ambiguity in the semantics-label graph; 3) ignoring the subtle semantic differences among the two tasks’ instances.
Co-guiding Net depicted in Sec. \ref{sec: coguid} is proposed to tackle the first two issues.
In this section, we focus on solving the third issue.
As stated in Sec. 1, the single-task and dual-task semantics contrastive relations can benefit single-task reasoning and dual-task mutual guidances.
Based on Co-guiding Net, we propose Co-guiding-SCL Net, which is augmented with our proposed single-task supervised contrastive learning and Co-guiding supervised contrastive learning, which are illustrated in Fig. \ref{fig: scl_model}.
The description of the notations in this section can be found in Table \ref{table: notation}.

\begin{table*}[t]
\centering
\fontsize{8}{10}\selectfont
\caption{Descriptions of notations used in Sec. \ref{sec: scl}.} 
\label{table: notation}
\setlength{\tabcolsep}{2mm}{
\begin{tabular}{l|l}
\toprule
Notation&Description \\ \midrule
$h^{[I, 0]}_u$ & Current sample's utterance representation in stage 1. $h^{[I, 0]}_u= \frac{1}{n}\sum^n_i h_i^{[I,0]}$ \\
$h_i^{[S,0]}$ & Current sample's $i$-th word representation of slot filling task in stage 1. \\
$h^{[I, 1]}_u$ & Current sample's utterance representation in stage 2. $h^{[I, 1]}_u = \frac{1}{n}\sum^n_i h_i^{[I,1]}$ \\
$h_i^{[S,1]}$ & Current sample's $i$-th word representation of slot filling task in stage 2. \\
$l_I$ & Current sample's intent label vector \\
$l^{j}_S$ & Current sample's $j$-th word's slot label vector\\
$l^{ss}$ & Current sample's sentence-level slot label vector\\
$l^{J}$ & Current sample's joint-task label vector for slot-guided multi-intent supervised contrastive learning.\\
$l^{J'}$ & Current sample's joint-task label vector for intent-guided slot supervised contrastive learning.\\

$K$ & The size of sample queues\\
$Q^0_u=\{h^k_{[uq,0]}\}_K^k$  &  Sample queue of the utterance representations in stage 1. \\
$h^k_{[uq,0]}$ & $k$-th instance in $Q^0_u$ \\
$Q^0_s=\{h^{[k,0]}_{[sq,0]}, ..., h^{[k,j]}_{[sq,0]}, ... h^{[k,n]}_{[sq,0]}\}^k_K$  &  Sample queue of the word representations of slot filling task in stage 1.  \\
$h^{[k,j]}_{[sq,0]}$ & $k$-th instance's $j$-th word representations in $Q^0_s$ \\
$Q^1_u=\{h^k_{[uq,1]}\}_K^k$  &  Sample queue of the utterance representations in stage 2. \\
$h^k_{[uq,1]}$ & $k$-th instance in $Q^1_u$ \\
$Q^1_s=\{h^{[k,0]}_{[sq,1]}, ..., h^{[k,j]}_{[sq,1]}, ... h^{[k,n]}_{[sq,1]}\}^k_K$  &  Sample queue of the word representations of slot filling task in stage 2.  \\
$h^{[k,j]}_{[sq,1]}$ & $k$-th instance's $j$-th word representations in $Q^1_s$ \\
$Q_l^I=\{l_k^I\}^k_K$  & Sample queue of multi-hot intent label vectors \\
$l_k^I$ & $k$-th instance's intent label vector \\
$Q_l^S = \{l^{[k,0]}_S, ..., l^{[k,j]}_S,..., l^{[k,n]}_S\}^k_K$&Sample queue of one-hot slot label vectors \\
$l^{[k,j]}_S$ & The one-hot slot label vector of the $k$-th instance's $j$-th word\\
$Q_l^ss=\{l_k^{ss}\}^k_K$  & Sample queue of sentence-level slot label vectors \\
$l_k^{ss}$ & $k$-th instance's sentence-level slot label vectors\\
$l^{J}_k$ & $k$-th instance's joint-task label vector for slot-guided multi-intent supervised contrastive learning.\\
$l^{J'}_k$ & $k$-th instance's joint-task label vector for intent-guided slot supervised contrastive learning.\\
  \bottomrule
\end{tabular}}
\end{table*}

Since our model performs supervised contrastive learnings, inspired by \cite{moco}, we maintain a set of sample queues that store not only the previously encoded features but also their labels: $Q^0_u, Q^0_s, Q^1_u, Q^1_s, Q_l^I, Q_l^S$ and $Q_l^{ss}$, whose descriptions can be found in Table \ref{table: notation}.
Note that the sentence-level slot label is used to provide sentence-level slot guidance in the proposed slot-guided intent supervised contrastive learning (Sec. 4.2.1).
Since it is not given in the datasets, we construct it by ourself and details can be found in Sec. 4.2.1.
After the current batch, we update the sample queues with the current batch’s features
and labels while dequeuing the oldest ones.

Next, we depict our proposed supervised contrastive learning mechanisms.

\subsection{Single-task Supervised Contrastive Learning}
In the first stage, initial estimation is performed to predict the initial labels that provide guidance for the other task.
The initial estimation is only based on the semantics of the current task.
As we state in Sec. \ref{sec:introduction}, there exist inherent semantics contrastive relations among the representations of each task.
Intuitively, the semantics representations corresponding to the same/similar intents or the same slots should be close to each other in the representation space.
Contrastively, the semantics representations corresponding to the same/similar intents or the same slots should be near to each other in the representation space.
To achieve this, we propose two supervised contrastive learning mechanisms for multiple intent detection and slot filling, respectively.
\subsubsection{Multi-Intent Supervised Contrastive Learning}
The function of this contrastive learning mechanism is to pull together the utterance representations that have the same/similar intent labels, while pushing apart the ones having different intent labels.
The anchor is $h^{[I, 0]}_u$ and the contrastive instances are from $Q^0_u$.
Unlike the conventional single-label multi-class supervised contrastive learning, our proposed multi-intent supervised contrastive learning can handle the fine-grained and dynamic relations among the multi-intent instances.
Specifically, this contrastive learning mechanism can be formulated as:
\begin{equation}
\begin{split}
\mathcal{L}_{\text{SCL}}^{\text{MI}} =& -\sum^{K}_k \mu_{k}\operatorname{log}\frac{{ e^{s(h^{[I, 0]}_u, h^k_{[uq,0]})}}}{\sum^{K}_j{e^{s(h^{[I, 0]}_u, h^j_{[uq,0]})}}}\\
\mu_{k} =& \frac{{l}_I\odot {l}_I^k}{\sum_j^K {l}_I\odot {l}_I^j}
\end{split}
\end{equation}
where $\odot$ denotes Hadamard product, $s(a,b) = \frac{a^{T}\cdot b}{\|a\|\cdot \|b\| \cdot \tau}$ is the cosine similarity function, and $\tau$ denotes contrastive learning temperature.
${l}_I\odot {l}_I^k$ denotes the golden similarity between the anchor and the $k$-th multi-intent instance.
A large ${l}_I\odot {l}_I^k$  denotes the $k$-th instance is quite similar to the anchor, leading to a large $\mu_{k}$ assigned to the loss function to pull them closer. 
Instead, if they have totally different labels, ${l}_I\odot {l}_I^k= 0$ and then $\mu_{k}=0$.
In this case, $s(h^{[I, 0]}_u, h^j_{[uq,0]})$, which denotes their distance, only appears in
the denominator.
As a result, the anchor and $k$-th multi-intent instance will be pushed apart by the negative gradient.

\subsubsection{Slot Supervised Contrastive Learning}
Since each word corresponds to only one slot label, this contrastive learning mechanism is conventional single-label multi-class supervised contrastive learning.
It aims to pull together the word representations that correspond to the same slot, while pushing apart the ones corresponding to different slots.
The anchor is $h^{[S,0]}_i$ and the contrastive instances are from $Q_s^0$.
Specifically, this contrastive learning mechanism can be formulated as:

\begin{equation}
\begin{split}
\mathcal{L}_{\text{SCL}}^{\text{S}} \!=\!& -\!\sum^{n}_i\sum^{n}_j\sum^{K}_k \frac{{l}_S^{i}\odot {l}_S^{[k,j]}}{M_i}\operatorname{log}\frac{{ e^{s(h^{[S, 0]}_i, h^{[k,j]}_{[sq,0]})}}}{E_i}\\
M_i =& \sum^n_j\sum^K_k {l}_S^{i} \odot {l}_S^{[k,j]}\\
E_i = &\sum^n_j \sum^{K}_j{e^{s(h^{[S, 0]}_i, h^{[k,j]}_{[sq,0]})}}
\end{split}
\end{equation}
${l}_S^{i}\odot {l}_S^{[k,j]}$ equals 1 or 0, indicating the $j$-th word representation of the $k$-th instance in $Q_s^0$ is the positive sample or negative sample of the $i$-th word representation of the current utterance.

\subsection{Co-guiding Supervised Contrastive Learning} \label{sec: coguiding scl}


In the second stage, the mutual guidances between the two tasks are achieved.
%
The representations (e.g., $h^{[I,1]_u}, h^{[S,1]_i}$) in the second stage contain two kinds of information: (1) the own task's semantics that can indicate the own task's labels; (2) the other task's initial label information that provides dual-task guidance.
Among the semantics representations of the two tasks, there exist dual-task semantics contrastive relations, which have been stated in Sec. \ref{sec:introduction}.
Therefore, we propose co-guiding supervised contrastive learning to integrate dual-task correlations into the contrastive learning procedure, which jointly considers both tasks' labels as the supervision signal to perform supervised contrastive learning.
Next, we introduce the details of slot-guided multi-intent supervised contrastive learning.

\subsubsection{Slot-guided Multi-Intent Supervised Contrastive Learning}
Multiple intent detection is a sentence-level classification task.
Although slot filling is word-level, the summarization of all of the slot labels in an utterance can provide sentence-level slot semantics.
For the utterances having the same intents, some of them may have different sentence-level slot semantics, which can be leveraged to discriminate the representations of these utterances.
And for the utterances having different intents, some of them may have similar sentence-level slot semantics, which can be leveraged to adjust the distances among their representations.
, learning better representations.
Slot-guided multi-intent supervised contrastive learning is proposed to achieve the above two aspects.

Firstly, we have to construct the sentence-level slot label by ourselves because it is not provided in the datasets.
The current utterance's sentence-level slot label vector is obtained by:
$
l^{ss} = \frac{\sum^n_{i,  {l}_S^i \neq \text{O}}  {l}_S^i}{\sum^N_{i = 1,  {l}_S^i \neq \text{O}}{1}}
$. 
The value of each dimension in $l^{ss}$ ranges from 0 to 1, which can be regarded as a score reflecting the degree of the corresponding slot for the sentence-level slot semantics.
Then we construct the joint-task label by concatenating $l_I$ with weighted $l^{ss}$: $l_{J}=\operatorname{concat}(l_I, \lambda^I * l^{ss})$, where $\lambda^I$ is a hyper-parameter.

Then the formulation of slot-guided multi-intent supervised contrastive learning is:
\begin{equation}
\begin{split}
\mathcal{L}_{\text{SCL}}^{\text{SGMI}} =& -\sum^{K}_k \mu_{k}\operatorname{log}\frac{{ e^{s(h^{[I, 1]}_u, h^k_{[uq,1]})}}}{\sum^{K}_j{e^{s(h^{[I, 1]}_u, h^j_{[uq,1]})}}}\\
\mu_{k} =& \frac{{l}^{J}\odot {l}_k^{J}}{\sum_j^K {l}^{J}\odot {l}^{J}_j}
\end{split}
\end{equation}
Note that ${l}^{J}\odot {l}_k^{J} = l_I \odot l_k^I + \lambda^I * \lambda^I * {l}^{ss}\odot {l}^{ss}_k$. In this way, $\lambda^I$ can control the extent the slot label is integrated for slot-guided multi-intent supervised contrastive learning.

\subsubsection{Intent-guided Slot Supervised Contrastive Learning}
Generally, the semantics of the intents expressed in an utterance is contained in each word's representation.
For some word representations from different utterances, their corresponding utterances may have different intents.
Even if they correspond to the same slot, their semantics are somehow different regarding the different intent semantics they contain.
And some other words' corresponding utterances may have the same/similar intents.
Even if they correspond to different slots, their semantics may be not quite different regarding the same/smilar intent semantics they contain.
The above two aspects can be leveraged to further discriminate the word representations corresponding to the same slot and adjust the distance among the ones corresponding to different slots.
To this end, we propose intent-guided slot supervised contrastive learning.

Firstly, we construct the joint-task label $l^{J'}_i=\operatorname{concat}(l^i_S, \lambda^S * l_I)$, where $\lambda^S$ is a hyper-parameter.
Then the intent-guided slot supervised contrastive learning can be formulated as:
\begin{equation}
\begin{split}
\mathcal{L}_{\text{SCL}}^{\text{IGS}} \!=\!& -\!\sum^{n}_i\sum^{n}_j\sum^{K}_k \frac{{l}^{J'}\odot {l}^{J'}_k}{M_i}\operatorname{log}\frac{{ e^{s(h^{[S, 1]}_i, h^{[k,j]}_{[sq,1]})}}}{E_i}\\
M_i =& \sum^n_j\sum^K_k {l}^{J'}_i \odot {l}^{J'}_{[k,j]} \\
E_i = &\sum^n_j \sum^{K}_j{e^{s(h^{[S, 1]}_i, h^{[k,j]}_{[sq,1]})}}
\end{split}
\end{equation}
Note that ${l}^{J'}_i\odot {l}^{J'}_{[k,j]} = l_S^i \odot l_S^{[k,j]} + \lambda^S * \lambda^S * {l}_I\odot {l}^{I}_k$. In this way, $\lambda^S$ can control the extent that intent labels are integrated for intent-guided slot supervised contrastive learning.

\subsection{Training Objective}
The final loss of Co-guiding-SCL Net is the sum of the loss of Co-guiding Net and all contrastive loss terms:
\begin{equation}
\begin{split}
\mathcal{L} = &\gamma\left(\mathcal{L}_I + \beta_I\mathcal{L}_I^{mp}\right) + \left(1-\gamma\right)\left(\mathcal{L}_S + \beta_S \mathcal{L}_S^{mp}\right) \\
            +& \eta_I\left(\mathcal{L}_{\text{SCL}}^{\text{MI}} + \mathcal{L}_{\text{SCL}}^{\text{SGMI}} \right) + \eta_S\left(\mathcal{L}_{\text{SCL}}^{\text{S}}+ \mathcal{L}_{\text{SCL}}^{\text{IGS}}\right)
\end{split}
\end{equation}
where $\eta_I$ and $\eta_S$ are hyper-parameters that balance the contrastive loss terms.
Note that all contrastive learning mechanisms only participate in the training process. 
Co-guiding Net and Co-guiding-SCL Net have the same inference procedure.

\section{Experiments}
\subsection{Datasets and Metrics}
Following previous works, MixATIS and MixSNIPS \cite{atis, snips,agif} are taken as testbeds.
MixATIS includes 13,162 utterances for training, 756 ones for validation and 828 ones for testing.
MixSNIPS includes 39,776 utterances for training, 2,198 ones for validation and 2,199 ones for testing.

As for evaluation metrics, following previous works, we adopt accuracy (Acc) for multiple intent detection, F1 score for slot filling, and overall accuracy (Acc) for the sentence-level semantic frame parsing.
Overall accuracy denotes the ratio of sentences whose intents and slots are all correctly predicted.

\begin{table}[b]
\centering
\fontsize{8}{10}\selectfont
\caption{The hyper-parameters tuned in our experiments.}
\setlength{\tabcolsep}{1.6mm}{
\begin{tabular}{c|c}
\toprule
word embedding dimension & 100, 128, 200, 256, 300 \\ \hline
label embedding dimension & 100, 128, 200, 256, 300  \\ \hline
hidden state dimension & 100, 128, 200, 256, 300  \\ \hline
layer number of GNNs & 2,3,4 \\ \hline
learning rate   &  5e-4, 1e-3, 5e-3 \\ \hline
weight decay   &  0, 1e-6 \\ \hline
$\beta_I$ & 1e-8, 1e-6, 1e-4, 1e-2, 1, 10 \\\hline
$\beta_S$ & 1e-8, 1e-6, 1e-4, 1e-2, 1, 10 \\\hline
$\tau$ & 0.05, 0.07, 0.1\\\hline
$\eta_I$ & 1e-4, 1e-3, 0.01, 0.1, 1 \\\hline
$\eta_S$ & 1e-4, 1e-3, 0.01, 0.1, 1 \\
\bottomrule
\end{tabular}}
\label{table: tuned_hyperparameter}
\end{table}

\begin{table*}[t]
\centering
\fontsize{8}{10}\selectfont
\caption{Results comparison. We report the average results of three runs with different random seeds. $\pm$ denotes Standard Deviation. $^\dag$ denotes our model significantly outperforms baselines with $p<0.01$ under t-test.} 
\setlength{\tabcolsep}{2mm}{
\begin{tabular}{l|ccc|ccc}
\toprule
\multirow{2}{*}{LSTM-based Models} & \multicolumn{3}{c|}{MixATIS} & \multicolumn{3}{c}{MixSNIPS} \\ \cline{2-7} 
                        & Overall(Acc)  &Slot (F1)  &Intent(Acc)& Overall(Acc)& Slot(F1)&Intent(Acc)           \\ \midrule
Attention BiRNN \cite{attbirnn} &  39.1 &  86.4      &   74.6      & 59.5        &  89.4     & 95.4 \\
Slot-Gated \cite{slot-gated}    &  35.5 &  87.7      &   63.9      & 55.4        &  87.9     & 94.6 \\
Bi-Model \cite{bimodel}         &  34.4 &  83.9      &   70.3      & 63.4        &  90.7    & 95.6 \\
SF-ID \cite{sfid}               &  34.9 &  87.4      &   66.2      & 59.9        &  90.6     & 95.0 \\
Stack-Propagation \cite{qin2019}&  40.1 &  87.8      &   72.1      & 72.9        &  94.2     & 96.0 \\
Joint Multiple ID-SF \cite{2019-joint}&36.1 &84.6    &   73.4      & 62.9        &  90.6     & 95.1 \\
AGIF \cite{agif}                &  40.8 &  86.7      &   74.4      & 74.2        &  94.2     & 95.1 \\
GL-GIN \cite{glgin}  &42.8 ($\pm$0.20) &87.9 ($\pm$0.36) &   76.0 ($\pm0.36$)     & 73.2 ($\pm0.42$)       &  93.9 ($\pm0.12$)     & 95.8 ($\pm0.31$) \\ \midrule
Co-guiding Net (ours)           &  \textbf{50.9}$^\dag$ ($\pm0.47$) &\textbf{89.5}$^\dag$ ($\pm0.49$) &\textbf{78.7}$^\dag$ ($\pm$0.32) & \textbf{77.2}$^\dag$ ($\pm$0.41) &  \textbf{94.9}$^\dag$ ($\pm$0.20) & \textbf{97.5}$^\dag$ ($\pm$0.16)\\
Co-guiding-SCL Net (ours)           &  \textbf{51.9}$^\dag$ ($\pm0.18$) &\textbf{90.0}$^\dag$ ($\pm0.14$) &\textbf{79.1}$^\dag$ ($\pm$0.18) & \textbf{77.1}$^\dag$ ($\pm$0.43) &  \textbf{95.0}$^\dag$ ($\pm$0.10) & \textbf{97.5}$^\dag$ ($\pm$0.14)\\
  \midrule  \midrule
\multirow{2}{*}{BERT-based Models} & \multicolumn{3}{c|}{MixATIS} & \multicolumn{3}{c}{MixSNIPS} \\ \cline{2-7} 
                        & Overall(Acc)  &Slot (F1)  &Intent(Acc)& Overall(Acc)& Slot(F1)&Intent(Acc)           \\ \midrule
BERT+Linear               &47.9 ($\pm$0.20) &87.0 ($\pm$0.35) &   80.6 ($\pm0.67$)     & 84.6($\pm0.48$)       &  96.8 ($\pm0.23$)     & 97.4 ($\pm0.30$) \\
BERT+GL-GIN \cite{glgin}  &49.3 ($\pm$0.82) &86.7 ($\pm$0.51) &   79.5 ($\pm0.21$)     & 84.9 ($\pm0.40$)       &  96.8 ($\pm0.18$)     & 96.9 ($\pm0.23$)\\ \midrule
BERT+Co-guiding Net (ours)  &  \textbf{52.4}$^\dag$ ($\pm0.32$) &\textbf{88.3}$^\dag$ ($\pm0.43$) &\textbf{82.3}$^\dag$ ($\pm$0.32) & \textbf{86.4}$^\dag$ ($\pm$0.21) &  \textbf{97.1}$^\dag$ ($\pm$0.18) & \textbf{97.4}$^\dag$ ($\pm$0.13)\\
BERT+Co-guiding-SCL Net (ours)  &  \textbf{54.0}$^\dag$ ($\pm0.44$) &\textbf{89.1}$^\dag$ ($\pm0.17$) &\textbf{84.2}$^\dag$ ($\pm$0.18) & \textbf{87.4}$^\dag$ ($\pm$0.12) &  \textbf{97.3}$^\dag$ ($\pm$0.02) & \textbf{98.2}$^\dag$ ($\pm$0.09)\\
  \midrule \midrule 
\multirow{2}{*}{RoBERTa-based Models} & \multicolumn{3}{c|}{MixATIS} & \multicolumn{3}{c}{MixSNIPS} \\ \cline{2-7} 
                        & Overall(Acc)  &Slot (F1)  &Intent(Acc)& Overall(Acc)& Slot(F1)&Intent(Acc)           \\ \midrule
RoBERTa+Linear               &48.4($\pm$0.32) &86.0 ($\pm$0.32) &   80.3 ($\pm0.49$)     & 82.1($\pm0.32$)       &  96.0 ($\pm0.24$)     & 97.4 ($\pm0.07$) \\
RoBERTa+GL-GIN \cite{glgin}  &49.9 ($\pm$0.35) &86.8 ($\pm$0.37) &   80.8($\pm0.19$)     & 82.5 ($\pm0.36$)       &  96.3 ($\pm0.64$)     & 97.3 ($\pm0.40$)\\ \midrule
RoBERTa+Co-guiding Net (ours)  &  \textbf{54.3}$^\dag$ ($\pm0.41$) &\textbf{88.4}$^\dag$ ($\pm0.35$) &\textbf{83.2}$^\dag$ ($\pm$0.30) & \textbf{83.9}$^\dag$ ($\pm$0.30) &  \textbf{97.5}$^\dag$ ($\pm$0.17) & \textbf{98.0}$^\dag$ ($\pm$0.20)\\
RoBERTa+Co-guiding-SCL Net (ours)  &  \textbf{56.6}$^\dag$ ($\pm0.43$) &\textbf{89.6}$^\dag$ ($\pm0.16$) &\textbf{84.4}$^\dag$ ($\pm$0.32) & \textbf{85.2}$^\dag$ ($\pm$0.07) &  \textbf{98.5}$^\dag$ ($\pm$0.06) & \textbf{98.3}$^\dag$ ($\pm$0.07)\\

    \midrule  \midrule
\multirow{2}{*}{XLNet-based Models} & \multicolumn{3}{c|}{MixATIS} & \multicolumn{3}{c}{MixSNIPS} \\ \cline{2-7} 
                        & Overall(Acc)  &Slot (F1)  &Intent(Acc)& Overall(Acc)& Slot(F1)&Intent(Acc)           \\ \midrule
XLNet+Linear           &52.5($\pm$0.24) &87.5 ($\pm$0.30) &   82.2 ($\pm0.43$)     & 84.3($\pm0.25$)       &  96.6 ($\pm0.20$)     & 97.2 ($\pm0.09$)\\
XLNet+GL-GIN \cite{glgin} &53.1($\pm$0.13) &87.8 ($\pm$0.24) &   82.6 ($\pm0.27$)     & 84.8($\pm0.28$)       &  96.6 ($\pm0.27$)     & 96.9 ($\pm0.09$) \\ \midrule
XLNet+Co-guiding Net (ours)   &  \textbf{54.0}$^\dag$ ($\pm0.20$) &\textbf{88.6}$^\dag$ ($\pm0.15$) &\textbf{83.8}$^\dag$ ($\pm$0.30) & \textbf{86.1}$^\dag$ ($\pm$0.18) &  \textbf{97.1}$^\dag$ ($\pm$0.13) & \textbf{97.6}$^\dag$ ($\pm$0.11)\\ 
XLNet+Co-guiding-SCL Net (ours)   &  \textbf{56.7}$^\dag$ ($\pm0.72$) &\textbf{89.2}$^\dag$ ($\pm0.21$) &\textbf{84.6}$^\dag$ ($\pm$0.28) & \textbf{87.7}$^\dag$ ($\pm$0.19) &  \textbf{97.6}$^\dag$ ($\pm$0.09) & \textbf{98.7}$^\dag$ ($\pm$0.09)\\
  \bottomrule
\end{tabular}}
\label{table: main results}
\end{table*}

\subsection{Implementation Details} \label{sec: implement}
We construct several group of our models, which are based on LSTM encoder and pre-trained language model (
PTLM) encoders (e.g. BERT \cite{bert}, RoBERTa \cite{roberta}, XLNet \cite{xlnet}).

\textbf{LSTM}.
Following previous works, the word and label embeddings are trained from scratch.
The dimensions of word embedding, label embedding, and hidden state are 256 on MixATIS, while on MixSNIPS they are 256, 128, and 256.
The layer number of all GNNs is 2.
Adam \cite{adam} is used to train our model with a learning rate of $1e^{-3}$ and a weight decay of $1e^{-6}$.
As for the coefficients Eq.\ref{eq: final objective}, $\gamma$ is 0.9 on MixATIS and 0.8 on MixSNIPS; on both datasets, $\beta_I$ is $1e^{-6}$ and $\beta_S$ is $1e^0$.
The above hyper-parameter settings are for both of Co-guiding Net and Co-guiding-SCL Net.
The hyper-parameters for the contrastive learning mechanisms in Co-guiding-SCL Net are set as follows.
$\tau$ is 0.07. $\eta_I$ and $\eta_S$ are 0.1 and 0.01.

\textbf{PTLM}.
The models based on PTLM encoders replace the self-attentive encoder with the PTLM encoder.
We adopt the base version for each PTLM encoder.
The learning rate is set to 1e-5 (tuning from [5e-6, 1e-5, 3e-5, 5e-5]) and adopt AdamW optimizer with default configuration.
Hidden state dimension is 768.
All other hyper-parameter setting are the same as LSTM-based model.

The model performing best on the dev set is selected then we report its results on the test set.
All experiments are conducted on RTX 6000 and DGX-A100 server.
\subsection{Baselines}
We compare our LSTM-based Co-guiding Net and Co-guiding-SCL Net with Attention BiRNN \cite{attbirnn}, Slot-Gated \cite{slot-gated}, SF-ID \cite{sfid}, Stack-Propagation \cite{qin2019}, Joint Multiple ID-SF \cite{2019-joint}, AGIF \cite{agif}  and GL-GIN \cite{glgin}.
We reproduce the results of GL-GIN using its official source code and default hyper-parameter setting.
We compare our PTLM-based Co-guiding Net and Co-guiding-SCL Net with PTLM-based GL-GIN, which is implemented by our self.
For fair comparison, we adopt the same learning rate and optimizer setting with our PTLM-based models. And other hyper-parameter setting are the same with GL-GIN.

\subsection{Main Results}
The performances of our models and baselines are shown in Table \ref{table: main results}, from which we have the following observations:
\subsubsection{Comparison of Our Models and Baselines}
(1) Co-guiding Net and Co-guiding-SCL Net gain significant and consistent improvements over baselines on all tasks and datasets.
Specifically, 
on MixATIS dataset, compared with GL-GIN, Co-guiding-SCL Net achieves significant improvements of 21.3\%, 2.4\%, and 4.1\% on sentence-level semantic frame parsing, slot filling, and multiple intent detection, respectively;
on MixSNIPS dataset, it overpasses GL-GIN by 5.3\%, 1.2\% and 1.8\% on sentence-level semantic frame parsing, slot filling and multiple intent detection, respectively.
The promising results of our model can be attributed to the mutual guidances between multiple intent detection and slot filling, allowing the two tasks to provide crucial clues for each other.
Besides, our designed HSLGs and HGATs can effectively model the interactions among the semantics nodes and label nodes, extracting the indicative clues from initial predictions.
And our proposed single-task supervised contrastive learning and co-guiding supervised contrastive learning can further capture single-task and dual-task semantics contrastive relations.

(2) Our models achieve larger improvements on multiple intent detection than slot filling.
The reason is that except for the guidance from multiple intent detection to slot filling, our models also achieve the guidance from slot filling to multiple intent detection, while previous models all ignore this.
Besides, previous methods model the semantics-label interactions by homogeneous graph and GAT, limiting the performance.
Differently, our model uses the heterogeneous semantics-label graphs to represent different relations among the semantic nodes and the label nodes, then applies the proposed HGATs over the graphs to achieve the interactions.
Consequently, their performances (especially on multiple intent detection) are significantly inferior to our model.

(3) The improvements in overall accuracy are much sharper.
We suppose the reason is that the achieved mutual guidances make the two tasks deeply coupled and allow them to stimulate each other using their initial predictions.
For each task, its final outputs are guided by its and another task's initial predictions.
By this means, the correct predictions of the two tasks can be better aligned.
As a result,  more test samples get correct sentence-level semantic frame parsing results, and then overall accuracy is boosted.

(4)
Based on PTLM encoders, our model brings more significant improvements than GL-GIN.
This is because GL-GIN performs semantic interactions, while PTLMs have strong abilities on semantics.
Differently, our models make the first attempt to achieve the semantics-label interactions, which cannot be achieved by PTLMs.
Therefore, our models' advantages do not overlap with PTLMs', and the high-quality semantics representations generated by PTLM can cooperate well with the co-guiding mechanism of our model.

\subsubsection{Comparison of Co-guiding Net and Co-guiding-SCL Net}
The performance improvements of Co-guiding-SCL Net over Co-guiding Net come from our proposed single-task supervised contrastive learning and co-guiding supervised contrastive learning.
And we can observe that Co-guiding-SCL Net can obtain larger improvements based on PTLM encoders than the LSTM encoder.
We suspect the reason is that PTLM can generate much higher-quality semantics representations than LSTM, and then the contrastive learning mechanisms can be improved because they are performed on the representations.

\begin{table*}[t]
\centering
\fontsize{9}{11}\selectfont
\caption{Results of ablation experiments.} 
\setlength{\tabcolsep}{2.5mm}{
\begin{tabular}{l|ccc|ccc}
\toprule
\multirow{2}{*}{Models} & \multicolumn{3}{c|}{MixATIS} & \multicolumn{3}{c}{MixSNIPS} \\ \cline{2-7} 
                        & Overall(Acc)  &Slot (F1)  &Intent(Acc)& Overall(Acc)& Slot(F1)&Intent(Acc)           \\ \midrule
Co-guiding Net          &  \textbf{50.9} &\textbf{89.5} &\textbf{78.7}
                         & \textbf{77.2}  &  \textbf{94.9} & \textbf{97.5} \\\midrule
w/o  S2I-guidance  &  47.4 ($\downarrow$3.5) &  88.5 ($\downarrow$1.0)      &   76.8    ($\downarrow$1.9) 
                & 76.3 ($\downarrow$0.9)       &  94.5 ($\downarrow$0.4)    & 96.7 ($\downarrow$0.8) \\
w/o  I2S-guidance &  47.3 ($\downarrow$3.6)&  88.4  ($\downarrow$1.1)    &   77.2  ($\downarrow$1.5)  
                 & 76.2    ($\downarrow$1.0)    &  94.7  ($\downarrow$0.2)  & 97.3 ($\downarrow$0.2)\\
w/o relations &  45.6 ($\downarrow$5.3) &  88.0  ($\downarrow$1.5)  & 77.5 ($\downarrow$1.2)   
                & 76.0       ($\downarrow$1.2) &  94.5    ($\downarrow$0.4) & 97.0 ($\downarrow$0.5) \\
+ Local Slot-aware GAT & 50.7  ($\downarrow$0.2) &  89.2  ($\downarrow$0.3)  &  78.6 ($\downarrow$0.1) 
                    & 75.6   ($\downarrow$1.6)     &  94.5 ($\downarrow$0.4)    & 96.1  ($\downarrow$1.4)  \\
  \bottomrule
\end{tabular}}
\label{table: ablation}
\end{table*}

\subsection{Model Analysis of Co-guiding Net}
We conduct a set of ablation experiments to verify the advantages of our work from different perspectives, and the results are shown in Table \ref{table: ablation}.
\subsubsection{Effect of Slot-to-Intent Guidance}
One of the core contributions of our work is achieving the mutual guidances between multiple intent detection and slot filling, while previous works only leverage the one-way message from intent to slot.
Therefore, compared with previous works, one of the advantages of our work is modeling the slot-to-intent guidance.
To verify this, we design a variant termed \textit{w/o S2I-guidance} and its result is shown in Table \ref{table: ablation}.
We can observe that Intent Acc drops by 2.0\% on MixATIS and 0.8\% on MixSNIPS.
Moreover, Overall Acc drops more significantly: 3.6\% on MixATIS and 0.9\% on MixSNIPS.
This proves that the guidance from slot to intent can effectively benefit multiple intent detection, and achieving the mutual guidances between the two tasks can significantly improve Overall Acc.

Besides, although both of \textit{w/o S2I-guidance} and GL-GIN only leverage the one-way message from intent to slot, \textit{w/o S2I-guidance} outperforms GL-GIN by large margins.
We attribute this to our proposed heterogeneous semantics-label graphs and heterogeneous graph attention networks, whose advantages are verified in Sec. \ref{sec: hslghgat}.

\subsubsection{Effect of Intent-to-Slot Guidance}
To verify the effectiveness of intent-to-slot guidance, we design a variant termed \textit{w/o I2S-guidance} and its result is shown in Table \ref{table: ablation}.
We can find that the intent-to-slot guidance has a significant impact on performance.
Specifically, \textit{w/o I2S-guidance} cause nearly the same extent of performance drop on Overall Acc, proving that both of the intent-to-slot guidance and slot-to-intent guidance are indispensable and achieving the mutual guidances can significantly boost the performance.

\subsubsection{Effect of HSLGs and HGATs} \label{sec: hslghgat}
In this paper, we design two HSLGs: (i.e., S2I-SLG, I2S-SLG) and two HGATs (i.e., S2I-HGAT, I2S-HGAT).
To verify their effectiveness, we design a variant termed \textit{w/o relations} by removing the relations on the two HSLGs.
In this case, S2I-SLG/I2S-SLG collapses to a homogeneous graph, and S2I-HGAT/I2S-HGAT collapses to a general GAT based on multi-head attentions.
From Table \ref{table: ablation}, we can observe that \textit{w/o relations} obtains dramatic drops on all metrics on both datasets. 
The apparent performance gap between \textit{w/o relations} and Co-guiding Net verifies that (1) our proposed HSLGs can effectively represent the different relations among the semantics nodes and label nodes, providing appropriate platforms for modeling the mutual guidances between the two tasks; (2) our proposed HGATs can sufficiently and effectively model interactions between the semantics and indicative label information via achieving the relation-specific attentive information aggregation on the HSLGs.

Besides, although \textit{w/o relations} obviously underperforms Co-guiding Net, it still significantly outperforms all baselines.
We attribute this to the fact that our model achieves the mutual guidances between the two tasks, which allows them to promote each other via cross-task correlations.

\subsubsection{Effect of I2S-HGAT for Capturing Local Slot Dependencies}
\cite{glgin} propose a Local Slot-aware GAT module to alleviate the uncoordinated slot problem (e.g., \textit{B-singer} followed by \textit{I-song}) \cite{slotrefine} caused by the non-autoregressive fashion of slot filling.
And the ablation study in \cite{glgin} proves that this module effectively improves the slot filling performance by modeling the local dependencies among slot hidden states.
In their model (GL-GIN), the local dependencies are modeled in both of the local slot-aware GAT and subsequent global intent-slot GAT.
We suppose the reason why GL-GIN needs the local Slot-aware GAT is that the global intent-slot GAT in GL-GIN cannot effectively capture the local slot dependencies.
GL-GIN's global slot-intent graph is homogeneous, and the GAT working on it treats the slot semantics nods and the intent label nodes equally without discrimination.
Therefore, each slot hidden state receives indiscriminate information from both of its local slot hidden states and all intent labels, making it confusing to capture the local slot dependencies.
In contrast, we believe our I2S-HLG and I2S-HGAT can effectively capture the slot local dependencies along the specific \textit{slot\_semantics\_dependencies} relation, which is modeled together with other relations.
Therefore, our Co-guiding Net does not include another module to capture the slot local dependencies.

To verify this, we design a variant termed \textit{+Local Slot-aware GAT}, which is implemented by augmenting Co-guiding Net with the Local Slot-aware GAT \cite{glgin} located after the Intent-aware BiLSTM$_s$ (the same position with GL-GIN).
And its result is shown in Table \ref{table: ablation}.
We can observe that not only the Local Slot-aware GAT does not bring improvement, it even causes performance drops.
This proves that our I2S-HGAT can effectively capture the local slot dependencies.

\subsection{Analysis of Supervised Contrastive Learning Mechanisms in Co-guiding-SCL Net}

\begin{figure}[t]
 \centering
 \includegraphics[width = 0.45\textwidth]{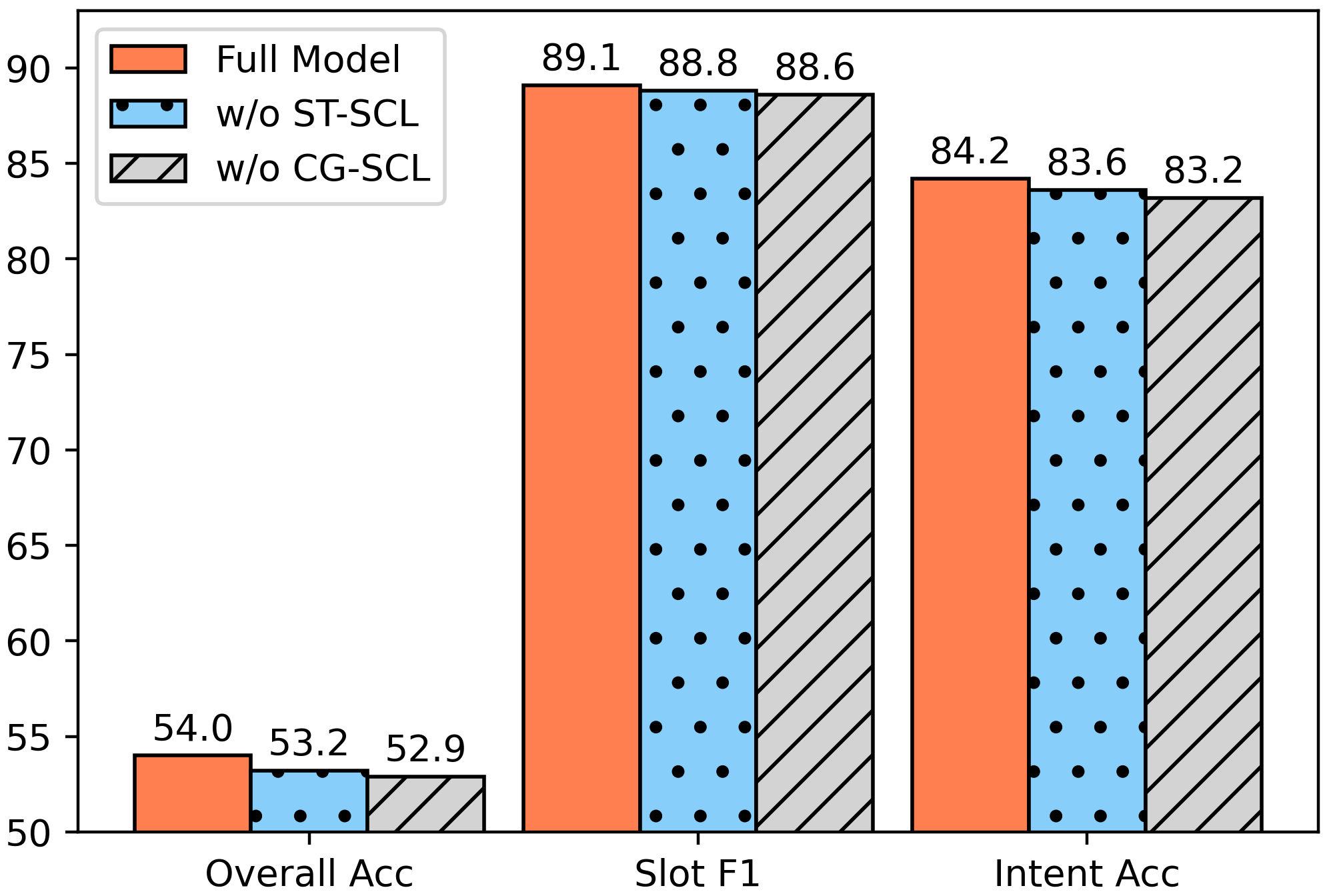}
 \caption{Ablation results. Full model denotes BERT+Co-guiding-SCL Net. w/o ST-SCL denotes the single-task supervised contrastive learning is removed. w/o CG-SCL denotes the co-guiding supervised contrastive learning is removed.}
 \label{fig: ablat_scl}
\end{figure}
\begin{figure}[t]
 \centering
 \includegraphics[width = 0.45\textwidth]{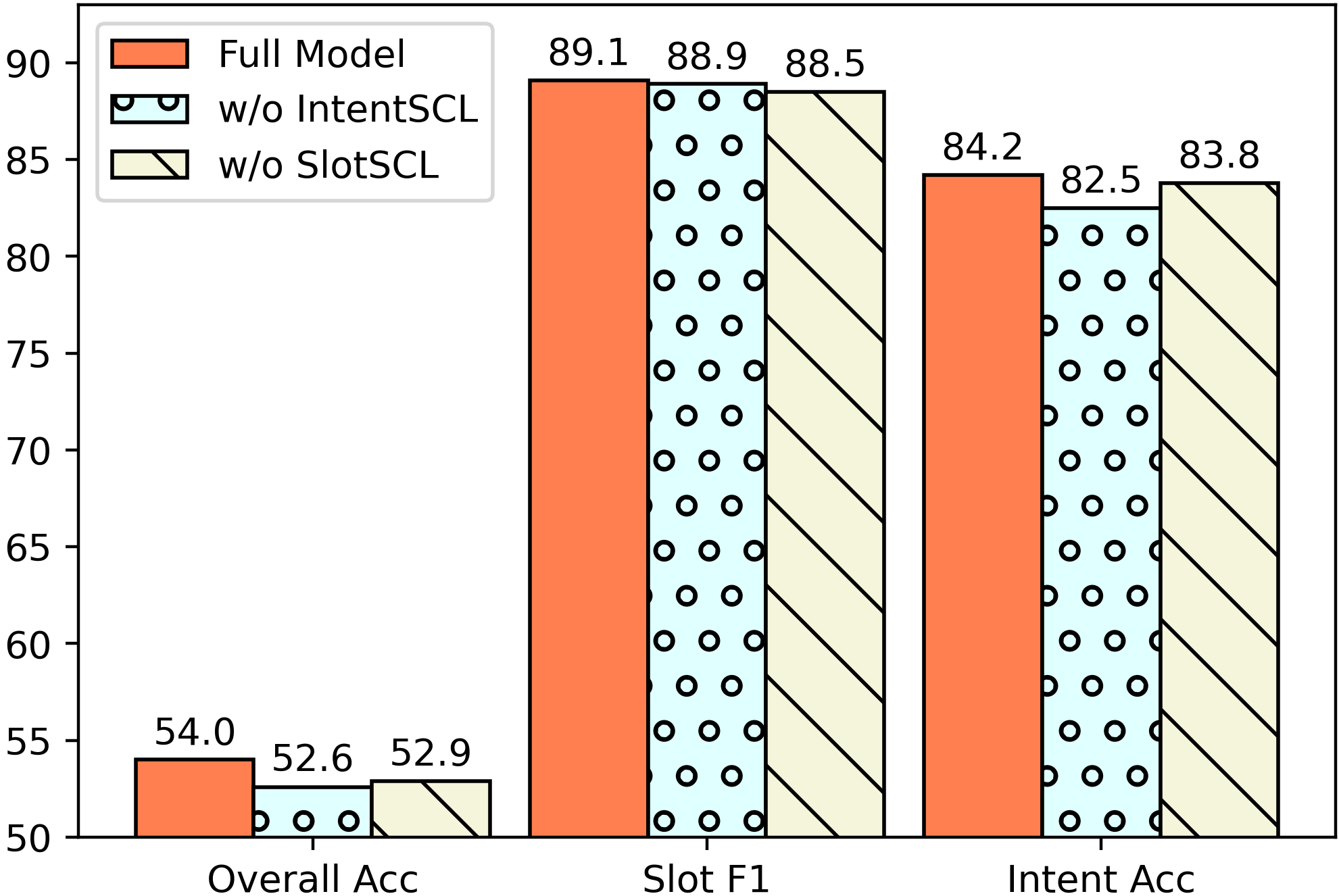}
 \caption{Ablation results. Full model denotes BERT+Co-guiding-SCL Net. w/o IntentSCL denotes the single-task and S=slot-guided multi-intent supervised contrastive learning mechanism are removed. w/o SlotSCL denotes the single-task and intent-guided slot supervised contrastive learning mechanisms are removed.}
 \label{fig: ablat_scl_loss}
\end{figure}

We conduct a set of ablation to verify the advantages of our proposed supervised contrastive learning mechanisms in Co-guiding-SCL Net, and the results on MixATIS dataset are shown in Fig. \ref{fig: ablat_scl} and \ref{fig: ablat_scl_loss}.

\subsubsection{Effect of Single-task/Co-guiding Supervised Contrastive Learning}
From Fig. \ref{fig: ablat_scl} we can observe that removing single-task supervised contrastive learning (ST-SCL) leads to performance decrease.
This is because ST-SCL can improve the label distribution of the initial estimations in the first stage via leveraging the single-task semantics contrastive relations, which is achieved by drawing together the representations corresponding to the same/similar labels while pushing apart the ones corresponding to different labels. 
And the better label distributions can provide more reliable indicative information for the dual-task co-guiding mechanism in the second stage.

We can find that the variant without co-guiding supervised contrastive learning (CG-SCL) perform significantly worse than the full model.
This proves the advantages of CG-SCL, which can further capture dual-task semantics contrastive relations at the second stage via integrating both tasks' labels as supervision signals for the supervised contrastive learning mechanism.
In the second stage, CG-SCL cooperates with the HGATs to comprehensively and effectively model the dual-task mutual guidances, significantly improving the final predictions.

\subsubsection{Effect of Intent/Slot Supervised Contrastive Learning}
From Fig. \ref{fig: ablat_scl_loss} we can observe that removing intent supervised contrastive learning (IntentSCL) leads to the performance decreases on intent accuracy, while causes the model performs worse on slot filling and sentence-level semantics parsing at the same time.
And removing slot supervised contrastive learning (SlotSCL) leads to the performance decreases not only on slot F1, but also on intent accuracy and overall accuracy.
There are two reasons.
First, IntentSCL and SlotSCL can effectively improve the model performances on multiple intent detection and slot filling, respectively.
Second, the co-guiding supervised contrastive learning further makes the two tasks deeply coupled and interrelated on each other's performances.
Therefore, removing anyone of IntentSCL and SlotSCL leads to the decreases on all of overall accuracy, slot F1 and intent accuracy.

\begin{figure*}[t]
 \centering
 \includegraphics[width = \textwidth]{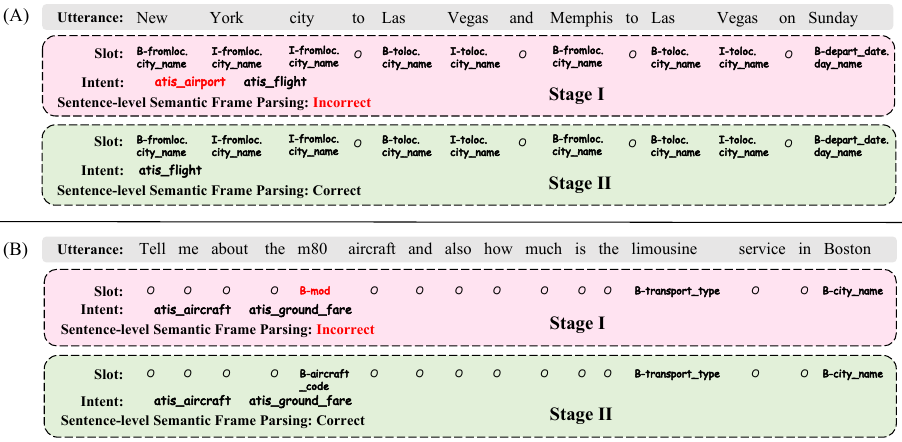}
 \caption{Case study of slot-to-intent guidance (A) and intent-to-slot guidance (B).  Red color denotes error.}
 \label{fig: case study}
\end{figure*}
\subsection{Case Study}

To demonstrate how our model allows the two tasks to guide each other, we present two cases in Fig. \ref{fig: case study}.

\subsubsection{Slot-to-Intent Guidance}
From Fig. \ref{fig: case study} (A), we can observe that in the first stage, all slots are correctly predicted, while multiple intent detection obtains a redundant intent \texttt{atis\_airport}.
In the second stage, our proposed S2I-HGAT operates on S2I-HLG.
It aggregates and analyzes the slot label information from the slot predictions of the first stage, extracting the indicative information that most slot labels are about \texttt{city\_name} while no information about \textit{airport} is mentioned.
Then this beneficial guidance information is passed into intent semantics nodes whose representations are then fed to the intent decoder for prediction.
In this way, the guidance from slot filling helps multiple intent detection predict correctly.

\subsubsection{Intent-to-Slot Guidance}
In the example shown in Fig. \ref{fig: case study} (B), in the first stage, correct intents are predicted, while there is an error in the predicted slots.
In the second stage, our proposed I2S-HGAT operates on I2S-HLG.
It comprehensively analyzes the indicative information of \textit{airecraft} from both of slot semantics node \textit{aircraft} and intent label node \texttt{atis\_aircraft}.
Then this beneficial guidance information is passed into the slot semantics of \textit{m80},
whose slot is therefore correctly inferred.

\subsection{Computation Efficiency}
\begin{table}[t]
\centering
\fontsize{8}{10}\selectfont
\caption{Comparison with SOTA on training time and latency. }
\setlength{\tabcolsep}{1.6mm}{
\begin{tabular}{c|c|c}
\toprule
Models  & \begin{tabular}[c]{@{}l@{}}Training Time\\ per Epoch \end{tabular}  &  \begin{tabular}[c]{@{}l@{}}Latency /Inference Time \\ per Utterance \end{tabular}  \\ \midrule
GL-GIN  &68s        & 2.6ms    \\ \hline
Co-guiding Net  &70s        & 2.9ms    \\ \hline
Co-guiding-SCL Net  &82s        & 2.9ms    \\ \hline\hline
BERT+GL-GIN  &148s        & 5.6ms    \\ \hline
BERT+Co-guiding Net  &156s        & 6.0ms    \\ \hline
BERT+Co-guiding-SCL Net  &185s        & 6.0ms    \\ \hline\hline
\end{tabular}}
\label{table: computing_efficient}
\end{table}
The training time and latency of our models and the state-of-the-art model are shown in Fig. \ref{table: computing_efficient}.
We can find that our Co-guiding-SCL Net costs some more training time due to the contrastive learning operations.
As for latency, both our Co-guiding Net and Co-guiding-SCL Net are comparable to GL-GIN, while they can significantly outperform it.
Our proposed contrastive learning mechanisms in Co-guiding-SCL only work in the training process, without affecting the latency.

\subsection{Zero-Shot Cross-Lingual Multi-Intent SLU} \label{sec: zero-shot}

\begin{table*}[t]
\centering
\fontsize{8}{9}\selectfont
\caption{Performances based on mBERT. $^*$ denotes our model significantly outperforms baselines with $p<0.05$ under the t-test. Best scores are in \textbf{bold}.} 
\setlength{\tabcolsep}{3mm}{
\begin{tabular}{l||c|c|c|c|c|c|c|c|c||c}
\toprule
\textbf{Intent Accuracy} & en & de & es & fr & hi & ja & pt & tr & zh & Avg.   \\ \hline
GL-CLEF &70.77&70.74&69.86&68.77&72.23&70.65&70.85&69.93&70.77&70.51\\
\midrule
Co-guiding Net (ours) &\textbf{97.65}&95.70&96.02&94.10&71.00&65.43&94.10&67.37&76.33 & 84.19\\
Co-guiding-SCL Net (ours)&97.54&96.52&96.56&94.69&75.92&64.56&94.21&72.07&76.19& 85.36\\
\midrule
GL-CLEF+Co-guiding Net (ours)&97.54&\textbf{96.82}&\textbf{97.25}&96.80&76.22&73.33&95.29&78.32&82.98&88.28\\
GL-CLEF+Co-guiding-SCL Net (ours)&97.54&96.52&96.72&\textbf{97.13}&\textbf{80.74}&\textbf{75.73}&\textbf{95.40}&\textbf{78.79}&\textbf{83.54}&\textbf{89.12}\\
\bottomrule
\toprule
\textbf{Slot F1} & en & de & es & fr & hi & ja & pt & tr & zh & Avg.   \\ \hline
GL-CLEF &94.85&85.77&85.51&84.99&56.52&65.92&81.08&65.66&78.47&77.64\\
\midrule
Co-guiding Net (ours)&96.28&81.57&83.63&81.38&41.06&39.06&74.00&51.93&64.21&68.12\\
Co-guiding-SCL Net (ours)&\textbf{96.30}&79.76&83.19&81.77&31.98&28.37&75.09&58.23&64.58&66.59\\\midrule
GL-CLEF+Co-guiding Net (ours)&95.84&84.64&85.34&84.80&58.62&\textbf{66.05}&\textbf{81.35}&66.84&\textbf{79.78}&78.14\\
GL-CLEF+Co-guiding-SCL Net (ours)&95.76&\textbf{86.92}&\textbf{86.12}&\textbf{85.86}&\textbf{59.87}&64.55&81.15&\textbf{68.92}&79.14&\textbf{78.70}\\
\bottomrule
\toprule
\textbf{Overall Accuracy} & en & de & es & fr & hi & ja & pt & tr & zh & Avg.   \\ \hline
GL-CLEF &66.55&48.69&46.37&45.98&18.92&\textbf{33.52}&45.85&23.26&42.40&41.28\\
\midrule
Co-guiding Net (ours)&\textbf{89.06}&56.88&57.15&52.89&9.63&8.80&46.11&15.57&28.59 &40.52 \\
Co-guiding-SCL Net (ours)&88.88&55.12&56.83&54.15&6.57&7.37&47.12&17.62&27.36&40.11\\\midrule
GL-CLEF+Co-guiding Net (ours)&88.24&64.87&62.07&61.27&20.34&27.58&59.30&29.09&\textbf{51.81}&51.62\\
GL-CLEF+Co-guiding-SCL Net (ours)&87.76&\textbf{67.90}&\textbf{63.30}&\textbf{62.16}&\textbf{22.62}&30.66&\textbf{59.94}&\textbf{31.89}&\textbf{51.81}&\textbf{53.11}\\
\bottomrule
\end{tabular}}
\label{table: mbert results}
\end{table*}

\begin{table*}[t]
\centering
\fontsize{8}{9}\selectfont
\caption{Performances based on XLM-R. $^*$ denotes our model significantly outperforms baselines with $p<0.05$ under the t-test. Best scores are in \textbf{bold}.} 
\setlength{\tabcolsep}{3mm}{
\begin{tabular}{l||c|c|c|c|c|c|c|c|c||c}
\toprule
\textbf{Intent Accuracy} & en & de & es & fr & hi & ja & pt & tr & zh & Avg.   \\ \hline
GL-CLEF &70.77&70.74&69.86&68.77&72.23&70.65&70.85&69.93&70.77&70.51\\
\midrule
Co-guiding Net (ours)&97.72&94.17&97.09&96.29&86.41&73.55&\textbf{96.79}&66.95&78.61      &87.51\\
Co-guiding-SCL Net (ours)&  97.80&94.58&97.29&95.20&84.06&68.92&96.41&55.99&83.91 &86.02\\\midrule
GL-CLEF+Co-guiding Net (ours)&  97.50&95.93&97.25&96.33&\textbf{87.64}&73.85&95.55&\textbf{75.76}&87.42 &\textbf{89.69}\\
GL-CLEF+Co-guiding-SCL Net (ours)&\textbf{98.17}&\textbf{96.15}&\textbf{97.83}&\textbf{97.01}&87.23&\textbf{77.35}&96.04&74.87&89.59&90.47\\
\bottomrule
\toprule
\textbf{Slot F1} & en & de & es & fr & hi & ja & pt & tr & zh & Avg.   \\ \hline
GL-CLEF &  96.01&85.30&86.85&83.00&63.95&67.92&79.99&56.24&\textbf{80.82}&77.78\\
\midrule
Co-guiding Net (ours)&  96.08&82.51&85.34&80.16&58.61&29.83&\textbf{80.33}&37.99&61.98    &68.09\\
Co-guiding-SCL Net (ours)&  \textbf{96.20}&83.95&85.50&81.60&63.70&32.68&80.04&42.29&57.31 &69.25\\\midrule
GL-CLEF+Co-guiding Net (ours)& 95.84&85.49&\textbf{87.29}&81.71&\textbf{72.05}&\textbf{72.22}&79.08&56.10&79.11 &\textbf{78.76}\\
GL-CLEF+Co-guiding-SCL Net (ours)&95.90&\textbf{86.59}&87.03&\textbf{81.84}&71.01&69.49&79.79&\textbf{56.79}&80.11&78.73\\
\bottomrule
\toprule
\textbf{Overall Accuracy} & en & de & es & fr & hi & ja & pt & tr & zh & Avg.   \\ \hline
GL-CLEF &66.48&48.28&46.78&45.85&22.28&28.48&42.30&15.94&44.53&40.10\\
\midrule
Co-guiding Net (ours)&88.43&60.54&61.50&54.07&23.33&5.94&\textbf{57.66}&5.13&22.36&42.11 \\
Co-guiding-SCL Net (ours)&  \textbf{88.69}&60.84&61.13&56.55&28.82&7.04&57.03&5.64&18.55 &42.70\\\midrule
GL-CLEF+Co-guiding Net (ours)& 87.98&64.24&63.59&58.20&\textbf{38.75}&\textbf{35.52}&55.87&19.81&53.83&53.09 \\
GL-CLEF+Co-guiding-SCL Net (ours)&88.50&\textbf{66.59}&\textbf{63.76}&\textbf{58.87}&35.83&34.73&56.99&\textbf{20.14}&\textbf{56.51}&\textbf{53.55}\\
\bottomrule
\end{tabular}}
\label{table: xlmr results}
\end{table*}

\subsubsection{Experiment Setup}

\noindent \textbf{Dataset and Metrics}
We evaluate our model on the multilingual benchmark dataset of MultiATIS++ \cite{multiatis++}.
This dataset includes the multi-intent training samples in English and the testing samples in 9 languages: English (en), Spanish (es), Portuguese (pt), German (de), French (fr), Chinese (zh), Japanese (ja), Hindi (hi) and Turkish (tr).
And intent accuracy (Acc), slot filling and overall accuracy (Acc) are adopted as the evaluation metrics.

\noindent \textbf{Baseline and Implementation}
Currently, the state-of-the-art model for zero-shot cross-lingual SLU is GL-CLEF \cite{gl-clef}, which utilizes the unsupervised contrastive learning to align the source language semantics and the target language semantics.
However, it is designed for single-intent SLU.
Therefore, we modify its official code to make it available for multi-intent SLU.
We use the sigmoid function and a linear layer, which is similar to Eq. 4, to replace its original intent classification module.
And we replace its original loss function with the loss function Eq. 11 used in our models. 

Apart from evaluating the performances of Co-guiding Net and Co-guiding-SCL Net,
we also combine them with GL-CLEF, forming GL-CLEF+Co-guiding Net and Co-guiding-SCL Net.
For fair comparison, the hyper-parameters of GL-CLEF used in GL-CLEF, GL-CLEF+Co-guiding Net and Co-guiding-SCL Net is directly retrieved from its original paper and the official code.
As for the hyper-parameters of  Co-guiding Net and  Co-guiding-SCL Net, we just use the ones staged in Sec. \ref{sec: implement}.

We conduct two groups of experiments, which are based on two multilingual pre-trained language models (e.g., mBERT \cite{bert} and XLM-R \cite{xlm-r}), respectively.
And we report the average results of three runs with different random seeds.
\subsubsection{Results Analysis}
The results of the models based on mBERT and XLM-R are shown in Table \ref{table: mbert results} and \ref{table: xlmr results}.
Firstly, we can observe that although Co-guiding Net and Co-guiding-SCL Net obtain significant improvements on English, it is hard to say that Co-guiding Net and Co-guiding-SCL Net can outperform GL-CLEF on the average overall accuracy of the total 9 languages.
We suspect the reason is that our models have a strong ability to model the mutual guidances between the two tasks and the semantics-label interactions, while there is no multilingual module in our models, which makes it hard to transfer the learned beneficial knowledge from the source language (English )to target languages.
However, if combining our models with GL-CLEF, we can observe obvious improvements.
Specifically, based on XLM-R, GL-CLEF+Co-guiding-SCL Net gains a relative improvement of 33.5\% over GL-CLEF on the average overall accuracy of the total 9 languages.
There are two reasons.
First, our models' advantage is capturing the beneficial knowledge learned from the source language via modeling the dual-task mutual guidances and capturing the fine-grained dual-task correlations.
Second, the semantics alignments of GL-CLEF work as a bridge that can transfer the knowledge learned from the source language to the target language.
Therefore, our models work well with GL-CLEF, achieving promising results for zero-shot cross-lingual multi-intent SLU.

\section{Conclusion}\label{sec: conclusion}
In this paper, 
we propose a novel two-stage framework that allows the two tasks to guide each other in the second stage using the predicted labels at the first stage.
Based on this framework, we propose two novel models: Co-guiding Net and Co-guiding-SCL Net.
To represent the relations among the semantics node and label nodes, we propose two heterogeneous semantics-label graphs and two heterogeneous graph attention networks to model the mutual guidances between intents and slots.
Besides, we propose the single-task supervised contrastive learning and co-guiding supervised contrastive learning, which are performed in the first stage and second stage, receptively.
We conduct extensive experiments to evaluate our models on multi-intent SLU and zero-shot cross-lingual multi-intent SLU.
Experiment results on benchmark datasets show that our model significantly outperforms previous models by large margins.
On multi-intent SLU task, our model obtains a relative improvement of 21.3\% over the previous best model on MixATIS dataset.
On zero-shot cross-lingual multi-intent SLU task, our model can relatively improve the state-of-the-art model by 33.5\%on average in terms of overall accuracy for the total 9 languages.

In addition, this work provides some general insights of exploiting the word-level and sentence-level semantics correlations via leveraging the dual-task contrastive relations among the word-level and sentence-level labels.
This idea can be applied to other scenarios which jointly tackle the sentence-level and word-level tasks.

\ifCLASSOPTIONcompsoc
  \section*{Acknowledgments}
\else
  \section*{Acknowledgment}
\fi

This work was supported by Australian Research Council Grant DP200101328.
Bowen Xing and Ivor W. Tsang was also supported by A$^*$STAR Centre for Frontier AI Research.

\ifCLASSOPTIONcaptionsoff
  \newpage
\fi

\normalem
\bibliographystyle{IEEEtran}
\bibliography{anthology.bib}
%
%
\begin{IEEEbiography}[{\includegraphics[width=1in,height=1.25in,clip,keepaspectratio]{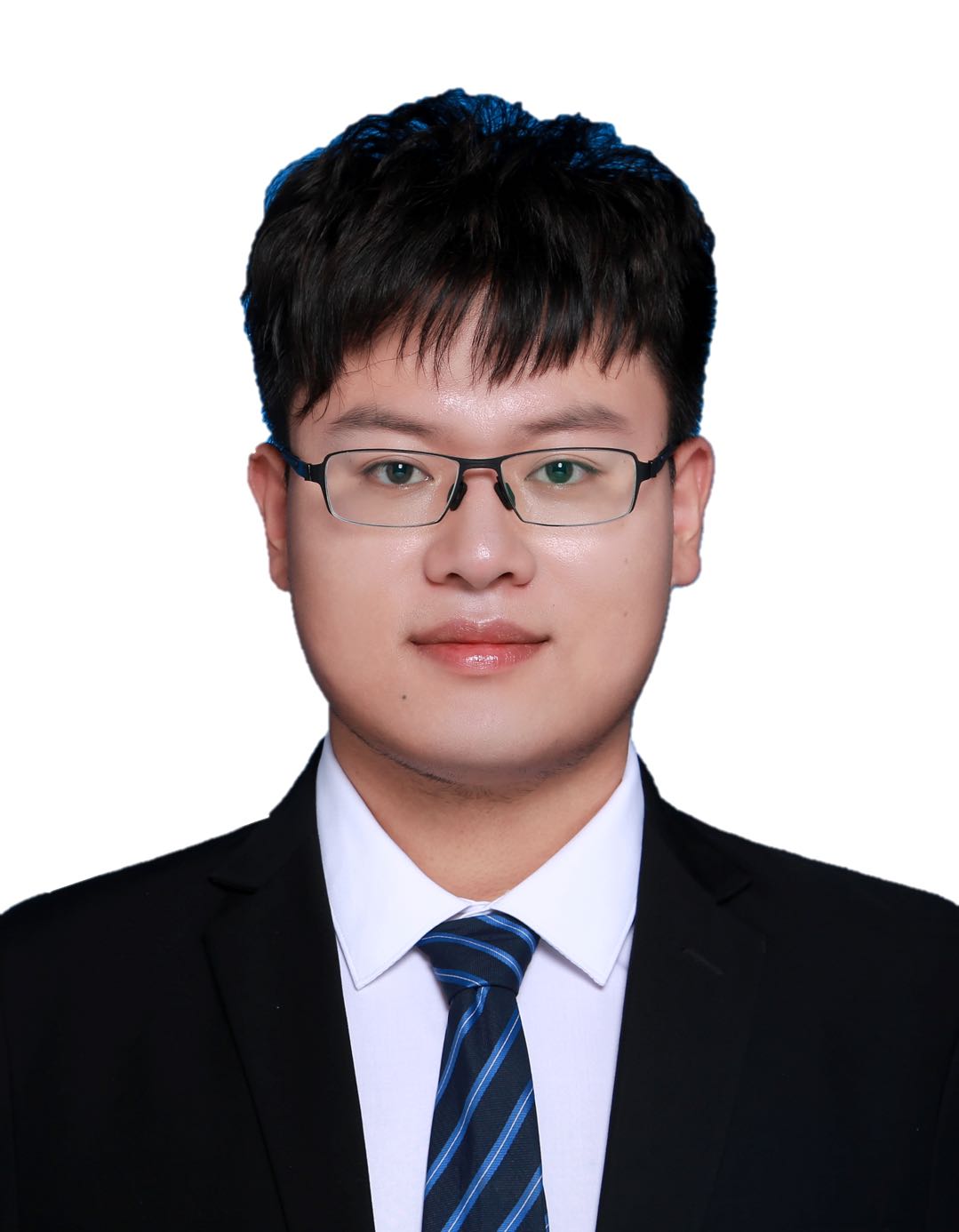}}]{Bowen Xing}
received his B.E. degree and Master degree from Beijing Institute of Technology, Beijing, China, in 2017 and 2020, respectively. He is currently a third-year Ph.D student at  the Australian Artificial Intelligence Institute (AAII), University of Technology Sydney (UTS).
His research focuses on graph neural networks, multi-task learning, sentiment analysis, and dialog system.
\end{IEEEbiography}

\begin{IEEEbiography}[{\includegraphics[width=1in,height=1.25in,clip,keepaspectratio]{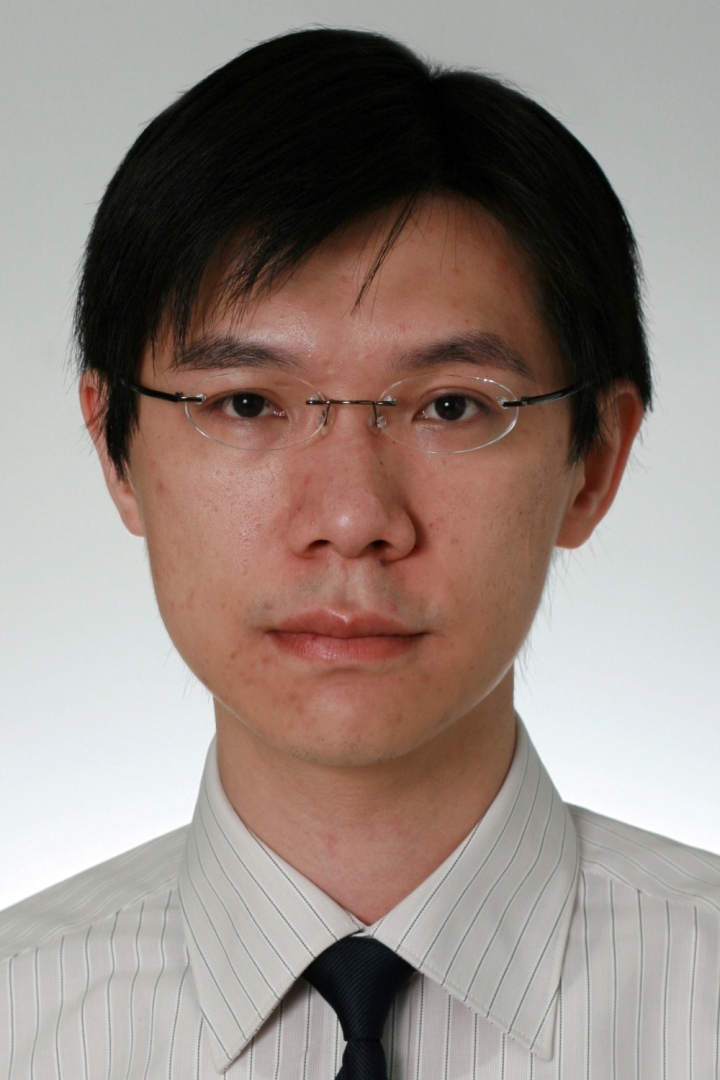}}]{Ivor W. Tsang} is an IEEE Fellow and the Director of A*STAR Centre for Frontier AI Research (CFAR). Previously, he was a Professor of Artificial Intelligence, at University of Technology Sydney (UTS), and Research Director of the Australian Artificial Intelligence Institute (AAII).
His research focuses on transfer learning, deep generative models, learning with weakly supervision, big data analytics for data with extremely high dimensions in features, samples and labels. His work is recognised internationally for its outstanding contributions to those fields.
In 2013, Prof Tsang received his ARC Future Fellowship for his outstanding research on big data analytics and large-scale machine learning. 
In 2019, his JMLR paper ``Towards ultrahigh dimensional feature selection for big data'' received the International Consortium of Chinese Mathematicians Best Paper Award. In 2020, he was recognized as the AI 2000 AAAI/IJCAI Most Influential Scholar in Australia for his outstanding contributions to the field, between 2009 and 2019. His research on transfer learning was awarded the Best Student Paper Award at CVPR 2010 and the 2014 IEEE TMM Prize Paper Award. In addition, he received the IEEE TNN Outstanding 2004 Paper Award in 2007 for his innovative work on solving the inverse problem of non-linear representations. Recently, Prof Tsang was conferred the IEEE Fellow for his outstanding contributions to large-scale machine learning and transfer learning.
Prof Tsang serves as the Editorial Board for the JMLR, MLJ, JAIR, IEEE TPAMI, IEEE TAI, IEEE TBD, and IEEE TETCI. He serves as a Senior Area Chair/Area Chair for NeurIPS, ICML, AAAI and IJCAI, and the steering committee of ACML.
\end{IEEEbiography}

\end{document}